## Title

Development and Validation of a Deep Learning Algorithm for Improving Gleason Scoring of Prostate Cancer

## Authors

Kunal Nagpal, MS[1], Davis Foote, BS[1], Yun Liu, PhD[1], Po-Hsuan (Cameron) Chen, PhD[1], Ellery Wulczyn, MS[1], Fraser Tan, PhD[1], Niels Olson, MD[2], Jenny L. Smith, DO[2], Arash Mohtashamian, MD[2], James H. Wren, MPH[3], Greg S. Corrado, PhD[1], Robert MacDonald, PhD[1], Lily H. Peng, MD, PhD[1], Mahul B. Amin, MD[4], Andrew J. Evans, MD, PhD[5], Ankur R. Sangoi, MD[6], Craig H. Mermel, MD, PhD[1], Jason D. Hipp, MD, PhD, FCAP[1], Martin C. Stumpe, PhD[1]

[1]Google AI Healthcare, Google, Mountain View, CA, USA

[2]Laboratory Department, Naval Medical Center San Diego, San Diego, CA, USA

[3]Henry M. Jackson Foundation

[4]Department of Pathology and Laboratory Medicine, University of Tennessee Health Science Center, Memphis, TN, USA

[5]Department of Pathology, Laboratory Medicine and Pathology, University Health Network and University of Toronto, Toronto, Ontario, Canada

[6]Department of Pathology, El Camino Hospital, Mountain View, CA, USA

## Corresponding author

Martin Stumpe, PhD, Google AI Healthcare, 1600 Amphitheatre Way, Mountain View, CA 94043 (mstumpe@google.com).

## Abstract

For prostate cancer patients, the Gleason score is one of the most important prognostic factors, potentially determining treatment independent of the stage. However, Gleason scoring is based on subjective microscopic examination of tumor morphology and suffers from poor reproducibility. Here we present a deep learning system (DLS) for Gleason scoring whole-slide images of prostatectomies. Our system was developed using 112 million pathologist-annotated image patches from 1,226 slides, and evaluated on an independent validation dataset of 331 slides, where the reference standard was established by genitourinary specialist pathologists. On the validation dataset, the mean accuracy among 29 general pathologists was 0.61. The DLS achieved a significantly higher diagnostic accuracy of 0.70 (p=0.002) and trended towards better patient risk stratification in correlations to clinical follow-up data. Our approach could improve the accuracy of Gleason scoring and subsequent therapy decisions, particularly where specialist expertise is unavailable. The DLS also goes beyond the current Gleason system to more finely characterize and quantitate tumor morphology, providing opportunities for refinement of the Gleason system itself.

## Introduction

Adenocarcinoma of the prostate is the second most common cancer diagnosed in men, with approximately one in nine men diagnosed in their lifetime.[1] For prostate cancer patients, subjective microscopic tissue examination remains the gold standard for diagnosis. Here, the Gleason score and tumor stage have remained the most powerful predictors of prognosis in virtually every large prostate cancer outcome study.[2] The Gleason system was initially developed in 1966 and stratifies prostate malignancies by tumor architectural patterns. The system has since been revised significantly[3,4] in an attempt to better reflect tumor biology. Importantly, the Gleason score (and its associated Gleason grade group[2]) is central to risk stratification and the National Comprehensive Cancer Network guidelines[5], which are widely-used clinically to guide standardized patient management decisions. Despite its indisputable role in prognostication and patient management, Gleason scoring by pathologists is a subjective exercise and suffers from suboptimal interobserver and intraobserver variability, with reported Gleason score discordance ranging from 30-53%.[6–14]

A potential approach to increasing the consistency and accuracy of Gleason grading lies in the field of artificial intelligence, where recent advances have been applied productively to imaging diagnostic tasks across dermatology[15,16], ophthalmology[17–20], radiology[21–23], and histopathology.[24–29] Artificial intelligence systems have been developed for prostate cancer detection in needle core biopsies[26] and Gleason grading of tissue microarrays[27], which comprise carefully selected sub-regions of tumor specimens used for research purposes, outside of routine clinical workflow. These studies have not explored the task of Gleason grading specimens used for clinical diagnosis.

Expertise and consistency in Gleason scoring have been shown to significantly improve its prognostic utility.[9,30] We thus reasoned that the availability of an accurate Gleason scoring tool for the whole-slide sections used in clinical workflows could help address the problem of grading variability, improve prognostication and optimize patient management. To this end, we developed a DLS to perform Gleason scoring and quantitation on prostatectomy specimens. The DLS accuracy is compared against a cohort of pathologists, where the reference standard was defined by genitourinary specialist pathologists. We further compared the risk stratification provided by our DLS, the cohort of pathologists, and our specialist-defined reference standard in predicting disease progression. Lastly, we also explored the potential of artificial intelligence to provide more fine-grained measures of tumor grading and the resulting potential to provide more precise prognostication.

## Results

**Overview of the Deep Learning System (DLS) and Data Acquisition.** Our approach is a 2-stage DLS: first a deep convolutional neural network-based regional Gleason pattern (GP) classification followed by a k-nearest-neighbor-based whole-slide Gleason grade group classification (Fig. 1). The first stage was trained using image patches extracted from the slide and the corresponding label derived from pathologist-labeled pixel-level annotations (Fig. 1). In total, we collected and used 112 million image patches derived from 912 slides (approximately 115,000 $mm^2$ of tissue), which required approximately 900 pathologist hours to annotate. To our knowledge, this constitutes the largest pixel-annotated histopathology dataset, roughly 4x larger in annotated tissue area than the training slides in the widely used Camelyon16 dataset[24]. The second stage was trained using 1,159 slide-level classifications provided by pathologists.

The DLS was evaluated on an independent validation dataset collected from three sources, consisting of 331 slides from 331 patients (Table 1). At least 3 pathologists provided initial reviews for each slide. A genitourinary specialist pathologist subsequently reviewed each slide along with the initial pathologists' comments to provide a final grade for use as the reference standard (Methods).

**Comparison of DLS to pathologists on Whole-Slide Gleason Scoring.** Independent of establishing the reference standard, we collected additional pathologist reviews on the validation dataset to compare with the DLS's performance. The mean accuracy among the 29 pathologists in classifying each slide's grade group was 0.61 (95% confidence interval (CI): 0.56-0.66). The DLS achieved an accuracy of 0.70 (95%CI 0.65-0.75), higher than the cohort of 29 ($p$=0.002, Fig. 2a). A subgroup of 10 pathologists in this cohort reviewed the entire validation dataset, with individual accuracies ranged from 0.53 to 0.73 (mean: 0.64). The DLS was more accurate than 8 of these 10 pathologists (Fig. 2b, Supplementary Table 4). The remaining 19 pathologists reviewed overlapping subsets of the validation set (see Methods), achieving individual accuracies ranging from 0.31 to 0.74 (mean: 0.60). Additional analyses are presented in Supplementary Tables 5-6 and Supplementary Fig. 1.

We additionally looked at three grade group (GG) decision thresholds: GG≥2, GG≥3, and GG≥4. The DLS achieved areas under receiver operating characteristic curves (AUCs) of 0.95-0.96 at each of these thresholds (Fig. 2c). The largest difference occurred at the GG≥4 threshold, where the DLS demonstrated both a higher sensitivity and specificity than 9 out of 10 individual pathologists.

**Comparison of DLS to pathologists on Gleason Pattern Quantitation.** In addition to the grade group, more granular reporting of the relative amounts of Gleason patterns is recommended by the International Society of Urological Pathology (ISUP), College of American Pathologists (CAP), World Health Organization (WHO), and recent publications.[31–34] As such, we also compared the DLS's accuracy in Gleason pattern quantitation to that of pathologists. Relative to the genitourinary pathologist reference standard, the DLS had a 4-6% lower mean absolute error than the average pathologist for quantitation of patterns 3 and 4 (Fig. 3). In subgroup analysis, for slides in grade groups 2 and 3 (where the amount of pattern 4 can change the overall grade group), the DLS again achieved better quantitation (8% lower mean absolute error). The trend for grade groups 4 and 5 (where quantitation of pattern 5 is significant) was similar. More details are available in Supplementary Tables 7-8.

**Insights from DLS Region-Level Classifications.** Furthermore, we evaluated the DLS's ability to classify tissue regions within each slide. We collected exhaustive region-level annotations for 79 slides, performed by 3 pathologists per slide, and compared the predictions of the DLS to these annotations (see Fig. 4a for an example). We first characterized the DLS's predictions by examining regions where the pathologists were concordant. For regions where all 3 pathologists agree on the same region classification (one of: non-tumor, Gleason pattern 3, 4, or 5), the DLS concurs 97% of the time. For the subset of these regions classified as a Gleason pattern, the DLS favors the same Gleason pattern as the pathologists 88% of the time (see Supplementary Results for an analysis of DLS errors).

Next, we characterized the DLS's prediction for regions where the pathologists were discordant by plotting the confidence score of the DLS for each category as a function of inter-pathologist agreement (Fig. 4b and Supplementary Fig. 2). For tissue regions where pathologists are concordant on Gleason pattern 3, discordant between 3 and 4, or concordant on Gleason pattern 4, the DLS' prediction scores change smoothly with the pathologists' classification distribution. The same trend is seen as we move from Gleason pattern 4 to 5. We further used the DLS's prediction scores directly to classify regions as *fine-grained Gleason patterns* (e.g. Gleason patterns 3.3 or 3.7). We found that by doing so, that DLS was able to represent a more gradual transition from well-to-poor differentiation than allowed by the canonical coarse Gleason pattern buckets (Fig. 4c and Supplementary Fig. 3).

**Measuring Effectiveness of Gleason Scoring in Risk Stratification for Disease Progression.** Lastly, we compared the ability of the DLS, the cohort of pathologists, and genitourinary specialist pathologists (who comprised the reference standard) to risk stratify patients for biochemical recurrence or disease progression (see Methods). In this analysis, we measured prognostic performance using the *c-index*, which is an extension of AUC that handles censored data in survival analysis. On the validation set, the DLS-predicted Gleason grade group achieved a c-index of 0.65. The pathologist-provided grade groups yielded a median c-index of 0.63 (see Methods), while the genitourinary specialist pathologists achieved a c-index of 0.69. Kaplan-Meier and hazard ratio analyses using a binary GG≥3 threshold, where hazard ratios for GG3 have previously been shown to be three-fold higher than GG2[2], to stratify patients into 'high risk' and 'low risk' categorizations showed the same trend (Fig. 5).

In addition to the risk stratification performance of GGs, we also used Cox models[35] to evaluate the prognostic ability of the underlying quantified Gleason patterns. The c-indices of these models were 0.697 for the DLS, 0.674 for the cohort of 29 pathologists, and 0.690 for the specialist-defined reference standard. As proof of concept that finer grained Gleason patterns can improve risk stratification, we also evaluated Cox-regression models trained on a more granular representation of the tumor pattern composition. Adding 'GP3.5' to the canonical Gleason patterns (thus summarizing the tumor composition as %GP3, %GP3.5, %GP4, and %GP5) raised the c-index to 0.704. Further adding %GP4.5 resulted in a c-index of 0.702 (Supplementary Table 10).

## Discussion

The present study shows that a DLS was more accurate than a cohort of 29 board-certified pathologists in Gleason scoring whole-slide images from prostatectomy patients.

The pathologists in this study had a 66% Gleason score concordance (61% Gleason grade group concordance) with genitourinary specialist pathologists, which is at the high end of several reported inter-pathologist Gleason score concordances of 47%-70%[6–14].

Previous studies have highlighted the value of expertise in pathologic interpretation. Central histologic reviews provided by pathologists experienced in genitourinary pathology improved prognostication relative to reviews provided by the local institution. Encouragingly, the risk stratification performance (as measured by the c-index and hazard ratio) in this study followed the same trend.[9,30] Due to the importance of genitourinary expertise in pathologic review, a second review has been recommended for high-risk patients after prostatectomy and for needle biopsies prior to prostatectomy.[8,9,36] In routine pathologic workflows, DLS-predicted Gleason scores could be computed on-demand and serve as a decision support tool. Future research is necessary to evaluate the potential clinical impact of the use of these predicted Gleason scores for patient prognostication and associated therapy decisions.

**Implications of DLS region-level pattern classifications and quantitation.** We further explored the implications of the DLS on each step of Gleason scoring and their respective scoring variability. The first aspect of Gleason scoring is the region-level classification of Gleason patterns across each slide. In this step, two-dimensional histologic examination of the three-dimensional tissue structures creates inherent ambiguity. Substantial additional variability arises from applying discrete categorizations to glandular differentiation that lies on a continuous spectrum, such as the Gleason pattern 3/4 transition between small glands and poorly defined acinar structures or the Gleason pattern 4/5 transition between fused glands and nests or cords.[12,37,38] Our data show that for regions where pathologists are discordant in Gleason pattern categorization, where the underlying histology is likely closer to the cusp between patterns, the DLS reflects this ambiguity in its prediction scores (Fig. 4b) and demonstrates the potential to assign finer-grained Gleason patterns (Fig. 4c). This finer-grained categorization provides opportunities to mitigate variability stemming from coarse categorization of a continuum, and opens avenues of research for more precise risk stratification (see Supplementary Table 10).

The next step in Gleason scoring after region-level categorization involves visual quantitation of the relative amounts of each Gleason pattern to determine the most prevalent patterns. Quantitation also allows for more granular prognostication. For example, prior studies have shown that prognosis of grade group 2-3 patients worsened for increases of percent Gleason pattern 4 as small as 5-10%.[34] As such, reporting of the quantitation of Gleason patterns is recommended.[4,31,39] However, visual quantitation is associated with inherent subjectivity.[40] In this regard, the DLS bypasses the variability introduced by visual quantitation through direct quantitation of Gleason patterns from its underlying region categorizations. The DLS's natural advantage in this regard and its more accurate quantitation than the cohort of pathologists (as measured by agreement with a specialist-adjudicated reference standard) suggest opportunity for more precise prognostication.

**Relation to previous works.** The above results complement previous works on the application of deep learning to prostate cancer histopathology. Campanella et al. demonstrated the use of deep learning in needle core biopsies to facilitate the detection of cancer foci.[26] Arvaniti et al. applied deep learning to Gleason score tissue microarrays.[27] This study complements prior work by applying deep learning to Gleason grading specimens that are more representative of a diversity of histologies and artifacts seen in routine clinical practice, and by directly comparing algorithmic performance with pathologists on a large multi-institutional dataset, with a rigorous reference standard defined by a team of board-certified pathologists and genitourinary specialist pathologists.

Another notable aspect of our work is the complexity and scale of the annotations required to develop our DLS. The complexity of Gleason grading has been discussed above; formalizing these interpretations as concrete annotations for training the DLS involved significant complexity, for example, "mixed" Gleason grades, artifacts, non-prostate tissue such as seminal vesicles, pre-malignant tissue, and uncommon variants. Please see Methods and Supplementary Methods for our detailed protocol. The size of this dataset was a key contributor to the accuracy of our DLS; training different models on titrated fractions of our dataset suggests that the DLS performance benefited greatly from the size of the dataset, and may yet improve with more or better quality data. Given the interobserver variability in Gleason grading, we also increased the accuracy of the pixel-level annotations in our tuning set by collecting triplicate annotations for each slide (see Methods and Supplementary Methods for details about the annotation and DLS training protocol).

In addition, our DLS stage-1 development process includes large scale, continuous "hard-negative mining" which aims to improve algorithm performance by running inference on the entire training dataset to isolate the hardest examples and further refine the algorithm using these examples. For histopathology applications on whole-slide imaging, this is a computationally expensive process, requiring inference over 112 million image patches in our training dataset. While previous works employing deep learning on histopathology images have employed hard negative mining in an offline "batch-mode"[24,41,42], we observed that performance improves with the frequency of inference on the entire training dataset, resulting in the "quasi-online" hard-negative mining approach (>30,000 DLS stage-1 inferences per second) used here. We anticipate that the benefits of this continuous hard negative mining approach may also be applicable to developing other histopathology deep learning algorithms.

**Limitations and future work.** This study has important limitations that would need to be addressed prior to implementation of associated tools in clinical practice. Unlike clinical environments, which are still largely based on glass slide review, this study focuses on digital review. Further, clinical environments enable pathologists to review additional sections, stains, or order consults for challenging cases. To account for this, pathologists were asked to indicate when they would prefer additional resources or consults to provide a more confident diagnosis. Corresponding sensitivity analysis excluding these cases is provided in Supplementary Table 9, showing qualitatively similar results.

Next, this study focuses on grading acinar prostatic adenocarcinoma (the vast majority of prostate cancer cases) in prostatectomy specimens, where the grade group informs postoperative treatment decisions rather than the decision to undergo the prostatectomy itself. As such, clinical outcomes after prostatectomy are less confounded by divergent treatment pathways than biopsies, supporting analyses of correlations with clinical follow-up data. In addition, prostatectomy specimens contain more tissue than biopsies, providing greater context during histological examination and improving the quality of the reference standard. However, important future work will generalize and validate the DLS for biopsies, other histologic variants, and other prognostic categorizations to aid clinical decisions throughout prostate cancer treatment pathways. Lastly, validation on larger clinically annotated datasets is required to evaluate the statistical significance of trends associated with prognostication demonstrated in this work.

## Conclusions

We have developed a DLS that outperforms a cohort of 29 generalist pathologists in Gleason scoring prostatectomy whole-slide images. Additionally, the DLS provides more accurate quantitation of Gleason patterns, finer-grained discretization of the well-to-poor differentiation spectrum, and opportunities for better risk stratification. In doing so, our DLS demonstrates the potential to enhance the clinical utility of the Gleason system for better treatment decisions for patients with prostatic adenocarcinoma.

## Methods

**Acquisition of Data.** De-identified, digitized whole-slide images of hematoxylin-and-eosin- (H&E-) stained formalin-fixed paraffin-embedded (FFPE) prostatectomy specimens were obtained from 3 sources: a public repository (The Cancer Genome Atlas, TCGA[43], n=397 patients), a large tertiary teaching hospital in the U.S. (n=361 patients), and an independent medical laboratory (n=11 patients, Table 1, Supplementary Table 1). This work was approved by the hospital's Institutional Review Board.

From TCGA we included all FFPE prostatectomy cases, the slides for which were scanned using a mix of scanners, including both Aperio and Hamamatsu scanners, and a mix of resolutions: ≈0.25µm/pixel ("40X magnification") and ≈0.5µm/pixel ("20X magnification"). From the hospital we included all prostatectomy cases where FFPE tissue blocks or slides were available based on a review of de-identified pathology notes. From the independent laboratory we obtained additional cases based on pathology reports to improve the representation of Gleason grade groups 4-5 in our study cohort (Table 1). From these sources, slides were obtained for cases within the 10-year Clinical Laboratory Improvement Amendments (CLIA) archival requirement, and tissue blocks for deaccessioned cases. Blocks were cut to produce sections of five-micron thickness and stained by CLIA-certified commercial laboratories (San Diego Pathology, San Diego, CA and Marin Medical Laboratories, Greenbrae, CA). Slides were digitized using a Leica Aperio AT2 scanner at a resolution of 0.25 µm/pixel.

Cases were randomly assigned to either the development (training/tuning) or independent validation datasets. For the 380 cases assigned to the validation dataset, pathologists identified one representative tumor-containing slide per case (see Grading section). Among these slides, 27 were excluded due to the presence of prostate cancer variants (Supplementary Table 2), 2 due to extensive artifacts or poor staining that hindered diagnosis, and 20 because of the inability of a genitourinary pathology specialist to confidently assign a diagnosis (Supplementary Table 3). The final validation dataset consisted of the remaining 331 slides (n=183 from TCGA, n=144 from the hospital, and n=4 from the laboratory).

**Overview of Pathologists' Annotations and Reviews.** A total of 35 pathologists reviewed slides for this study, all of whom completed residency in human anatomical pathology. Twenty-nine pathologists were U.S.-board-certified (the "cohort of 29"), and another 3 had genitourinary specialization (1 Canadian-board-certified and 2 U.S.-board-certified). The remaining 3 pathologists were formerly board-certified or certified outside of North America, and provided annotations for the training and tuning datasets but not the validation dataset.

We collected slide-level reviews and region-level annotations from pathologists. Slide-level reviews categorize each slide into its Gleason grade group. Region-level annotations label specific tissue regions (such as specific Gleason patterns) within a slide. We describe the annotation protocol for the validation dataset here, and include additional details and the protocol for the training and tuning datasets in the "Grading" section and Supplementary Figure 5 in the Supplement.

**Collection of Slide-Level Reference Standard.** The slide-level reference standard was used to validate the DLS's and general pathologists' performance. For each slide, the reference standard was provided by one genitourinary specialist pathologist. To improve accuracy, the specialist reviewing each slide also had access to initial Gleason pattern percentage estimates and free-text comments from prior reviews of at least 3 general pathologists. The specialist then determined the final GP percentages for tumor of each Gleason Pattern (GP): %GP3, %GP4, and %GP5 for use as the reference standard. We derived the slide-level Gleason score and corresponding grade group (1, 2, 3, or 4-5) based on the predominant and next-most-common

Gleason patterns provided by the genitourinary specialist, avoiding variability introduced by inconsistent application of "tertiary replacement" (see "Grading" in the Supplement). All slides were reviewed in a manner consistent with ISUP 2014 and CAP guidelines with no time constraint.[4,31]

**Collection of Slide-Level Reviews for Pathologists' Performance.** To evaluate general pathologists' performance at Gleason scoring, we collected additional slide-level reviews for each slide, independent from those collected for determining the reference standard. These reviews came from a total of 29 pathologists. From this cohort, 10 pathologists provided reviews for every slide in the validation dataset. The remaining 19 pathologists reviewed overlapping subsets of the validation set (median: 53 slides, range: 41-64), collectively providing three reviews per slide.

**Collection of Region-Level Annotations.** To compare region-level DLS predictions to pathologist interpretations, pathologists provided annotations of specific tissue regions within a slide, outlining individual glands or regions and providing an associated label (non-tumor, or GP 3, 4, or 5). For these time-consuming region-level annotations, a subset of the validation dataset (79 of 331 slides) were selected based on slide-level grade group diversity. Each of these 79 slides was exhaustively annotated by three pathologists (≥95% tissue coverage; taking on average 3 hours per pathologist per slide). Only regions for which all three pathologists provided a label were used for validation.

**Clinical Follow-up Data.** To measure risk stratification performance, we used additional clinical follow-up data. For the TCGA subset of data, we used the progression free interval as the clinical endpoint, as recommended by the authors of the TCGA Clinical Data Resource.[44] For the hospital subset, biochemical recurrence, as defined by a postoperative prostate specific antigen (PSA) measurement of $\geq$ 0.4[45], was used as the clinical endpoint. Clinical endpoints were not available from the medical laboratory and for a small number of cases from TCGA and the hospital. Of the 331 validation slides, 320 had available clinical follow-up data.

**Deep Learning System (DLS).**The DLS consists of 2 stages (Fig. 1), which correspond to the region-level annotations and slide-level reviews: first a regional classification, and subsequent whole-slide Gleason grade group classification. The first stage segments each slide into small image patches and feeds each patch into a convolutional neural network that classifies each patch as one of four classes: non-tumor, or Gleason pattern 3, 4, or 5. When applied to the entire whole-slide image, this stage outputs a "heatmap" indicating the categorization of each patch in the tissue section. The second stage consists of a nearest-neighbor classifier that uses a summary of the heatmap output from the first stage to categorize the grade group of each slide. We briefly outline the DLS development procedure below, and provide additional details in the "Deep Learning System" section in the Supplement.

The first stage's convolutional neural network is a modified InceptionV3[46], that classifies each tissue region of roughly 32×32μm by using input image patches of 911×911μm centered on the region. The label for each region was derived from the pathologist-provided region-level annotations (see Supplementary Methods, "Grading" section). Ensembling and hard-negative mining were employed to further improve model performance (see Supplementary Methods, "Hard-negative Mining" section).

In the second stage of the DLS, we first obtained a categorical prediction for each region by taking the class with the highest calibrated likelihood, where calibration weights were determined empirically using the tuning set. Next, for each slide, the number of regions predicted as each category was summarized and used for evaluation of (GP) quantitation (%GP3, %GP4, and %GP5). The three %GPs, together with the tumor involvement, were used as features (Fig. 1), similar to what a pathologist would need for Gleason scoring. Finally, we trained k-nearest neighbor classifiers for several prediction tasks: four-class grade group (GG) classification (1, 2, 3 or 4-5), and each of three binary classifications of GG≥2, GG≥3, and GG≥4.

**Statistical Analysis.** We assessed the DLS's Gleason scoring performance relative to the reference standard for slide-level and region-level classifications. For slide-level grade group categorization, we compared the accuracy of the DLS to the mean of the 29 individual pathologist accuracies, where accuracy is the fraction of exact matches with the reference standard. This provided equal representation of each pathologist despite their differing number of reviews. We additionally measured performance using accuracy adjusted by a population-level grade group distribution[47], and Cohen's kappa[48]. For the three binary classifications of slide-level grade group, we used the area under the receiver operating characteristic curve (AUC). For quantitation of relative Gleason patterns in the tumors, we computed the mean absolute error (MAE).

For clinical follow-up analysis, the concordance index was used to measure the overall effectiveness of grade group risk-stratification with respect to an adverse clinical endpoint (disease progression or biochemical recurrence as described above). The hazard ratio and associated Kaplan-Meier curves were used to evaluate risk-stratification at the binary classification of GG ≥ 3. For these risk stratification analyses, the cohort-of-29 pathologists Grade Group classifications were sampled to approximate equal representation of each pathologist (see "Statistical Analysis" in the Supplement). Analysis on the sampled classifications that produced the median concordance and hazard ratios respectively among 999 sampling iterations is reported here.

Confidence intervals for all evaluation metrics were computed using a bootstrap approach (see “Statistical Analysis” in the Supplement). All statistical tests were 2-sided permutation tests. A p-value<0.05 was considered statistically significant. No adjustment for multiple comparisons was made. These analyses were performed in Python (v2.7.6), using the scikit-learn (v0.19.1) and lifelines (v0.12.0) libraries.

**Reporting Summary.** Further information on experimental design is available in the Nature Research Reporting Summary linked to this article.

**Code availability.**

Code Availability. The deep learning framework used here (TensorFlow) is available at https://www.tensorflow.org/. The Python libraries used for computation and plotting of the performance metrics (SciPy, NumPy, Lifelines, and MatPlotLib) are available under https://www.scipy.org/, http://www.numpy.org/, https://github.com/CamDavidsonPilon/lifelines/, and https://matplotlib.org/, respectively.

**Data availability.**

The dataset from TCGA that was used in this study is available from the Genomic Data Commons portal (https://portal.gdc.cancer.gov/), which is based upon data generated by the TCGA Research Network (http://cancergenome.nih.gov/). The other datasets are not publicly available due to restrictions in the data sharing agreements with the data sources. The use of de-identified tissue for this study was approved by an Institutional Review Board (IRB).

## Figures/Tables

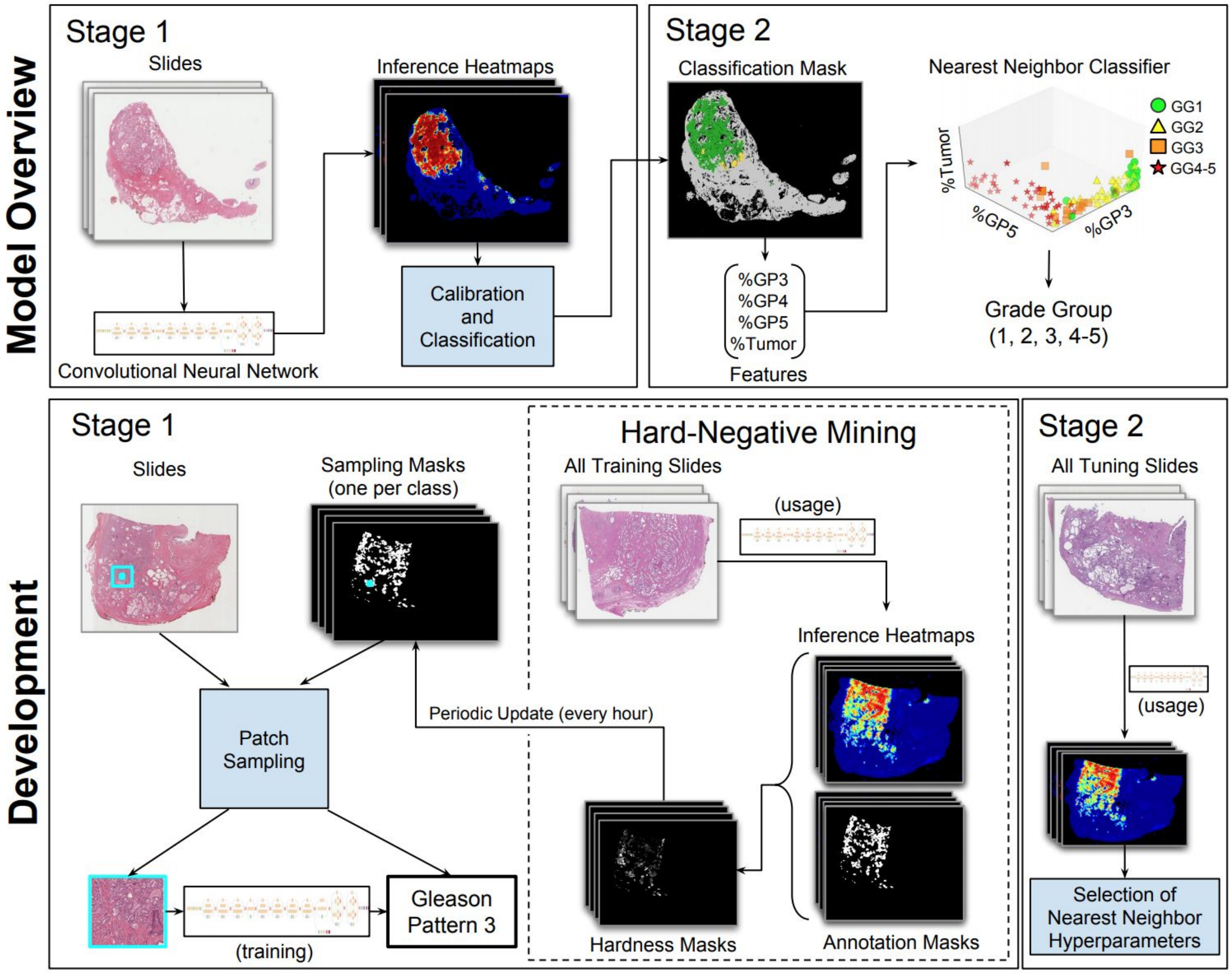

**Fig. 1 | Illustration of the development and usage of the two-stage deep learning system (DLS).** Developing the DLS involves training two machine learning models. Stage 1 is an ensembled deep convolutional neural network (CNN) that classifies every region in the slide as non-tumor, or its Gleason pattern (GP). Training the stage 1 CNN involves first collecting pathologists' annotations (Annotation Masks) of whole-slide images at the region level, and then generating "sampling masks" indicating the locations of each of the four classes (non-tumor, GP3, GP4, and GP5) for each slide. Over the course of millions of training iterations, sampled image patches and associated labels are used to train the constituent CNNs in the ensembled stage 1 CNN model. During the training process, we performed hard-negative mining by periodically applying each individual partially trained model to the entire training corpus of whole-slide images. Comparison of these intermediate inference results to the original annotations highlights the most difficult image patches, and we focus training on these patches. Stage 2 involves first collecting pathologists' labels of the Gleason grade group (GG) for each slide. Next, the predictions of the stage 1 model are calibrated and converted to 4 features that indicate the amount of tumor and each GP in the slide. k-nearest neighbor (kNN) classifiers are then trained to predict the GG (1, 2, 3, or 4-5), or whether the GG is above specific thresholds

(GG≥2, GG≥3, or GG≥4). For more details, please refer to the “Deep Learning System” section in the Supplement.

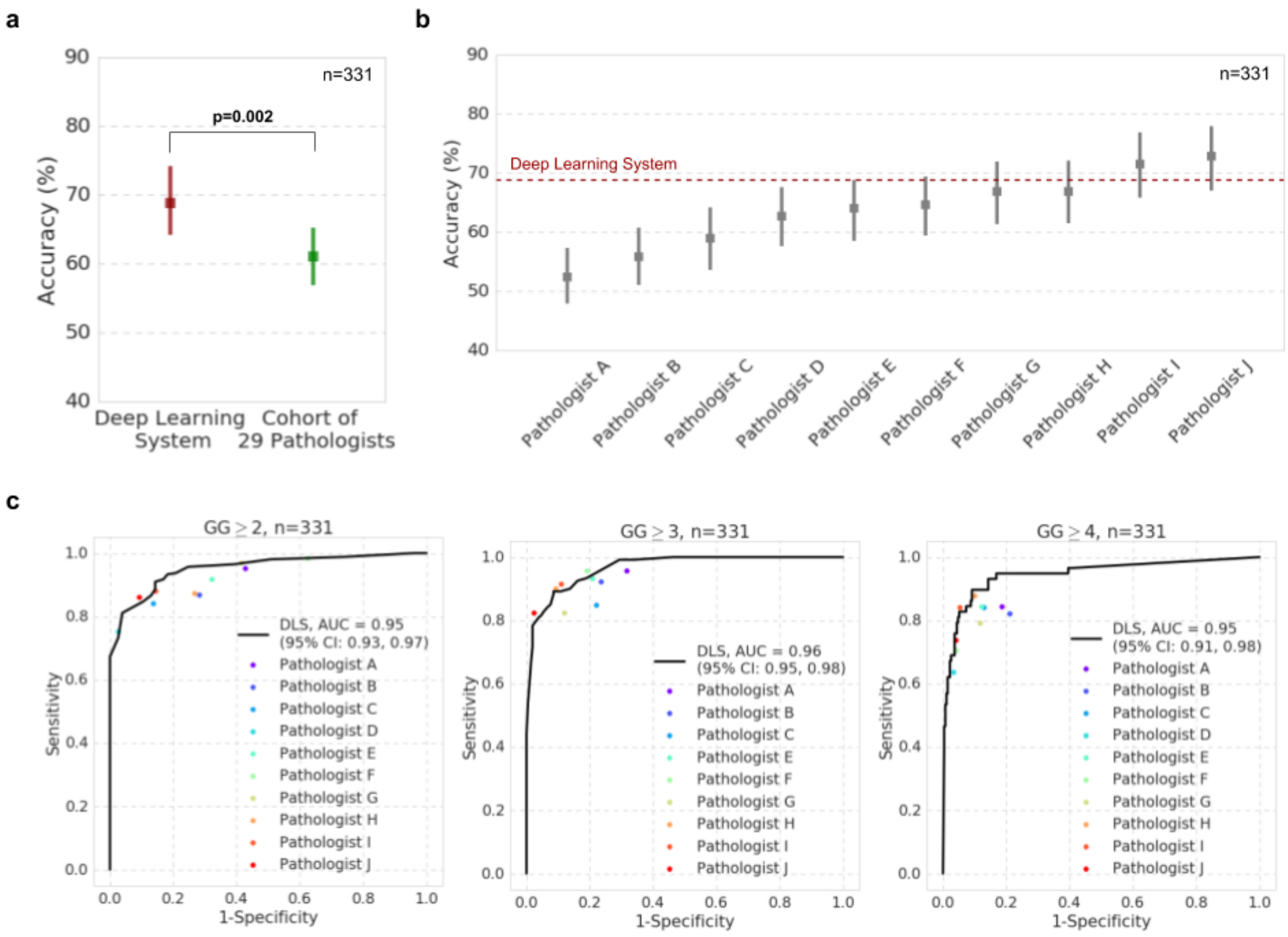

**Fig. 2 | Comparison of prostate cancer Gleason scoring performance of the deep learning system (DLS) with pathologists. a,** Accuracy of the DLS (in red) compared with the mean accuracy among a cohort-of-29 pathologists (in green). Error bars indicate 95% confidence intervals. **b,** Accuracy of the DLS compared to 10 individual pathologists (among the cohort of 29, indicated by pathologists A-J) who reviewed all of the slides in the validation set. See eTable 4 in the Supplement for more details. **c,** The receiver operating characteristic curves compare the sensitivity and specificity of the DLS with individual pathologists and the cohort-of-29 pathologists for binary classification of whether the Gleason grade group (GG) is above the thresholds of GG≥2, GG≥3 and GG≥4. AUCs and associated 95% confidence intervals for the DLS are provided in the legend. Higher and to the left indicates better performance.

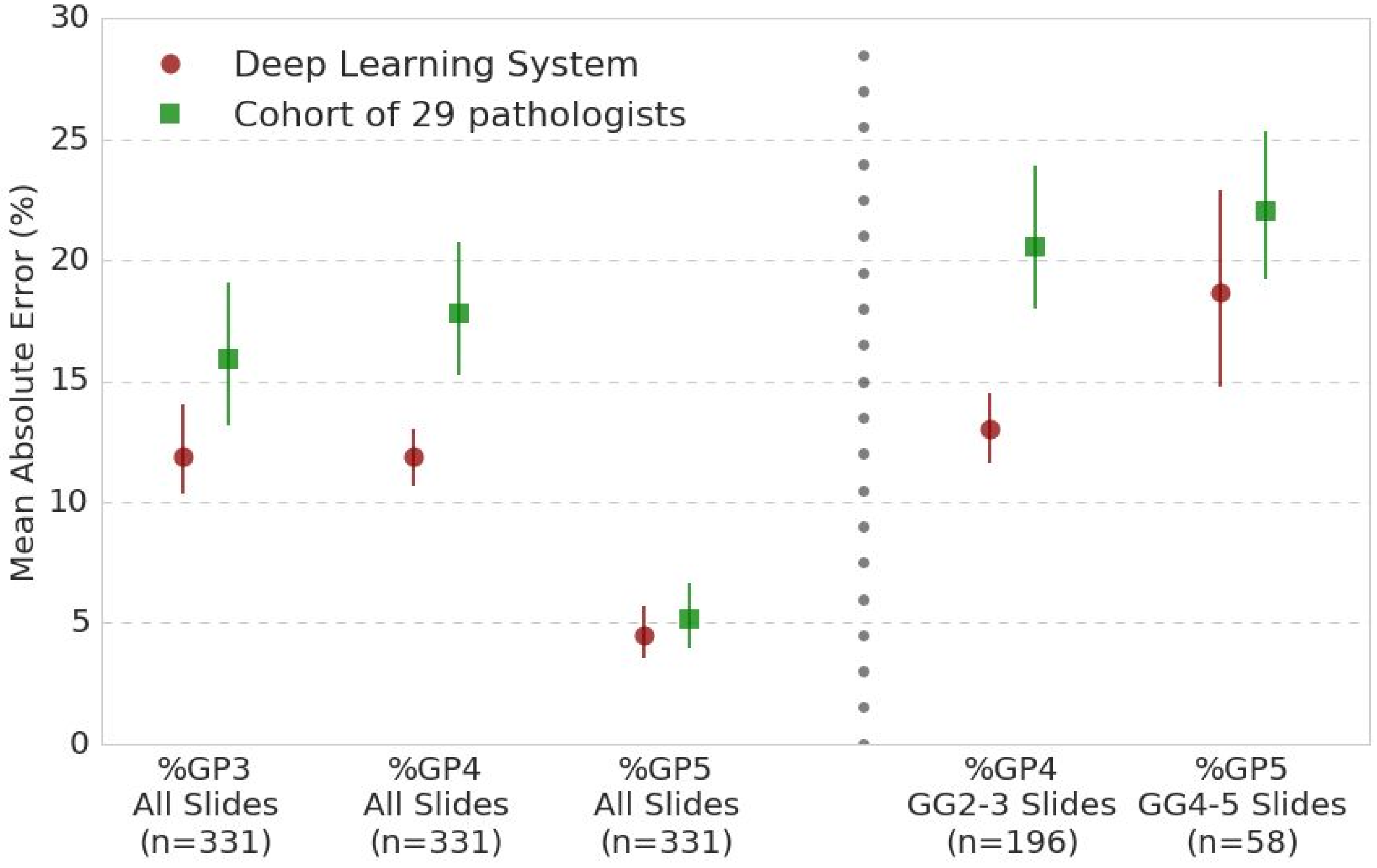

**Fig. 3 | Comparison of the deep learning system (DLS) with pathologists for Gleason Pattern (GP) quantitation.** Each dot indicates the mean average error (lower is better) for Gleason pattern quantitation, with error bars show the 95% confidence intervals. Left: overall Gleason pattern quantification results among all slides. Right: subgroup analysis where Gleason pattern quantification is of particular importance: grade group 2-3 slides where percent of Gleason pattern 4 can change the overall grade group, and grade group 4-5 slides where percent of Gleason pattern 5 reporting is recommended by the College of American Pathologists.

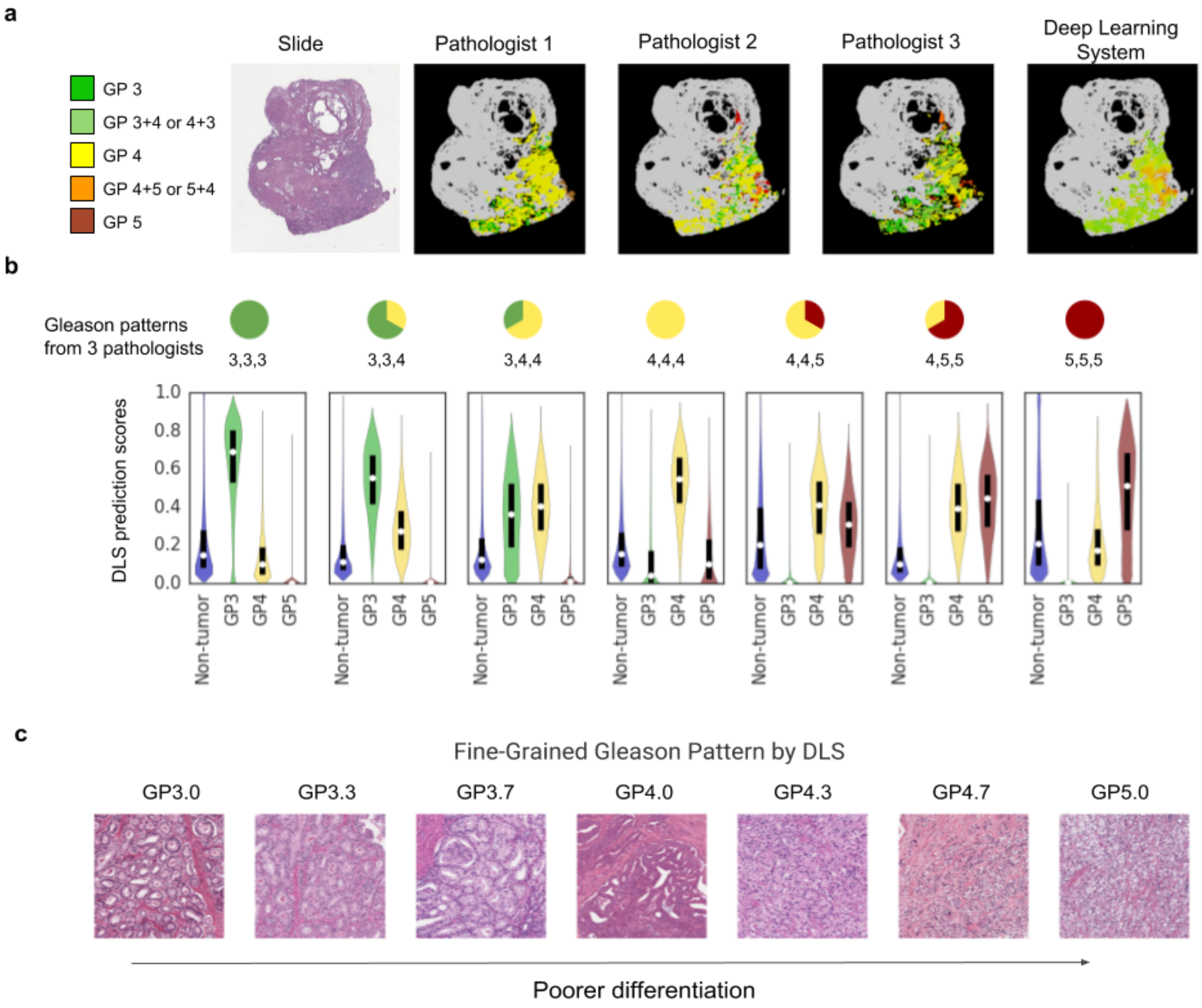

**Fig. 4 | Assessing the region-level classification of the DLS. a,** 3 pathologists annotated this slide with general concordance on the localization of tumor areas, but poor agreement on the associated Gleason patterns: a 'pure' grade like Gleason pattern 3, 4, or 5, or a mixed grade comprising features of more than one pure pattern. The DLS assigned each image patch to a fine-grained Gleason pattern, as illustrated by the colors interpolating between Gleason patterns 3 (green), 4 (yellow), and 5 (red). See the "Fine-grained Gleason Pattern" section in the Supplement. **b,** Quantification of the observations from panel A across 79 slides (41 million annotated image patches) for which 3 pathologists exhaustively categorized every slide. The violin plots indicate DLS prediction-likelihood distributions. The white dots and black bars identify medians and interquartile ranges, respectively. The predicted likelihood of each Gleason pattern by the DLS changes smoothly with the pathologists' classification distribution. See Supplementary Fig. 2 for a similar analysis on images with mixed-grade labels. **c,** The continuum of Gleason patterns learned by the DLS reveals finer categorization of the well-to-poorly differentiated spectrum (see "Fine-grained Gleason Pattern" section in the Supplement). Each displayed image region is the region closest (of millions in our validation dataset) to its labeled quantitative Gleason pattern. Columns 1, 4, and 7 represent regions for

which the highest confidence predictions are Gleason patterns 3, 4, and 5 respectively. The columns in between represent quantitative Gleason patterns between these defined categories. See Supplementary Fig. 3 for additional examples.

**a**

| Grade Group Classifications | C-Index for Adverse Clinical Endpoint (95%CI) |
|---|---|
| Cohort-of-29 Pathologists | 0.63 (0.52-0.73) |
| Deep Learning System | 0.65 (0.54-0.76) |
| Specialist-defined reference standard | 0.69 (0.58-0.78) |

**b**

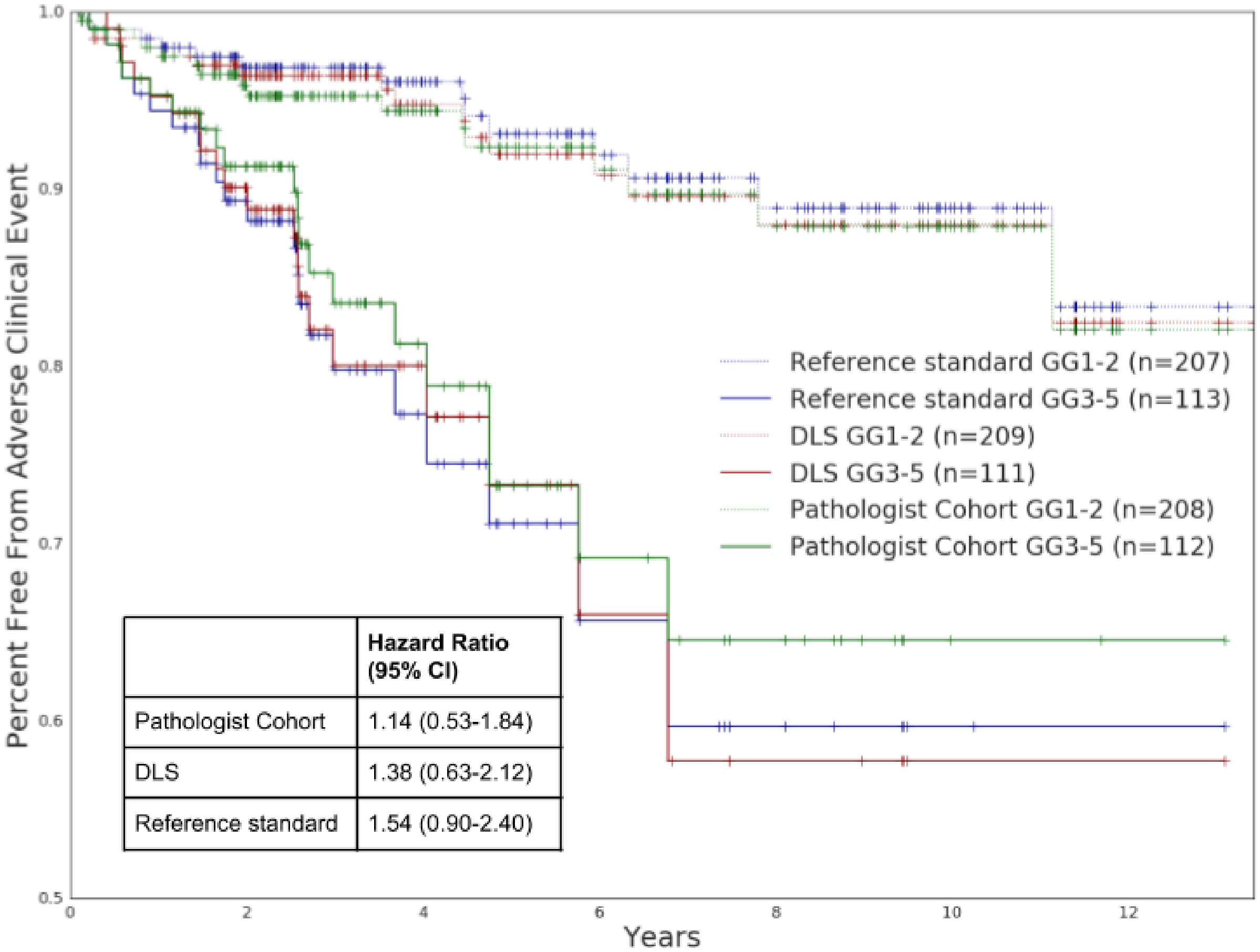

| | Hazard Ratio (95% CI) |
|---|---|
| Pathologist Cohort | 1.14 (0.53-1.84) |
| DLS | 1.38 (0.63-2.12) |
| Reference standard | 1.54 (0.90-2.40) |

**Fig. 5 | Comparison of risk stratification between pathologists, deep learning system, and the specialist-defined reference standard. a,** Concordance index provided by each entity's grade group (GG) classification (GGs 1, 2, 3, 4-5) in stratisfying adverse clinical endpoints of disease progression or biochemical recurrence (BCR) (see "Clinical Follow-up Data" in Methods). 95% confidence intervals were obtained by bootstrapping. For the cohort-of-29 pathologists, the median c-index is reported (See "Statistical Analysis" in Supplementary Methods). **b,** Kaplan-Meier curves using a binary threshold (GG$\geq$3) for risk stratification. Dotted lines correspond to the lower risk group (GG1-2) and solid lines correspond to the higher risk group (GG3-5). A larger separation between the risk groups indicates better risk stratification. Tick marks indicate censorship events. For the cohort-of-29 pathologists, analyses of sampled Grade group classifications that produced a median hazard ratio are plotted here (See "Statistical Analysis" in Supplementary Methods).

| | Source or Diagnosis | TCGA | Tertiary Teaching Hospital | Medical Laboratory | Total |
|---|---|---|---|---|---|
| **Number of Patients** | Patients with available prostatectomy specimens | 219 | 157 | 4 | 380 |
| | Excluded due to non-gradable prostate cancer variants | 22 | 5 | 0 | 27 |
| | Excluded due to extensive image artifacts or poor staining | 2 | 0 | 0 | 2 |
| | Specialist unable to provide confident diagnosis | 12 | 8 | 0 | 20 |
| **Slide-level Gleason grade group** | Patients in study (1 slide per patient) | 183 | 144 | 4 | 331 (100%) |
| | Grade group 1 | 10 | 67 | 0 | 77 (23%) |
| | Grade group 2 | 77 | 57 | 0 | 134 (40%) |
| | Grade group 3 | 46 | 14 | 2 | 62 (19%) |
| | Grade group 4-5 | 50 | 6 | 2 | 58 (18%) |
| | Grade group 4 | 10 | 2 | 0 | 12 (4%) |
| | Grade group 5 | 40 | 4 | 2 | 46 (14%) |
| **Region-level Gleason pattern annotations** | Number of slides | 62 | 14 | 3 | 79 |
| | Non-tumor (patches) | 18,022,643 | 10,879,735 | 2,152,853 | 31,055,231 |
| | Gleason pattern 3 (patches) | 2,445,437 | 343,685 | 2,016 | 2,791,138 |
| | Gleason pattern 4 (patches) | 4,288,977 | 8,280 | 106,227 | 4,403,484 |
| | Gleason pattern 5 (patches) | 1,797,331 | 326 | 129,059 | 1,926,716 |

**Table 1: Number and breakdown of slides in the validation dataset.** The validation set contains prostatectomy cases from three sources. A representative slide was selected from each patient's case. The reference standard for the Gleason scores in the validation set was established by an initial review by at least 3 pathologists from a cohort of 19 and then

adjudication by 1 of 3 genitourinary specialists. The low prevalence of grade groups 4 and 5 in our dataset prompted us to merge these two groups for more reliable statistical comparisons.

**Acknowledgements**

For technical advice and discussion, we thank the following, who are all employees of Alphabet Inc: Tim Hesterberg, PhD, Michael Howell, MD, MPH, David Miller, MS, Alvin Rajkomar, MD. For software infrastructure, logistical support, and slide digitization services, we thank members of the Google AI Healthcare Pathology team. Lastly, we are deeply grateful to the pathologists who provided annotations for this study.

**Author Contributions**

K.N., D.F., Y.L., F.T., J.D.H., M.C.S. designed the experiments; K.N., D.F., Y.L., C.C., E.W., J.W. wrote code to achieve different tasks; F.T., N.O., J.S., A.M., J.W., L.P., R.M., G.S.C., C.H.M. acquired the tissue samples for use in the study and provided strategic support; M.B.A., A.J.E., A.R.S. provided labels for use in measuring algorithm performance; M.S.C. supervised the project; K.N., D.F., Y.L., C.C., J.D.H. wrote the manuscript with the assistance and feedback of all other co-authors.

**Competing interests**

K.N., D.F., Y.L., C.C., E.W., F.T., G.S.C., R.M., L.P., C.H.M., J.D.H., and M.S.C. are employees of Google Inc. and own Alphabet stock.

**Disclaimer**

The views expressed in this article are those of the author(s) and do not necessarily reflect the official policy or position of the Department of the Navy, Department of Defense, nor the U.S. Government. A.M., N.O., J.L.S., and J.H.W. are military Service members. This work was prepared as part of their official duties. Title 17, U.S.C., §105 provides that copyright protection under this title is not available for any work of the U.S. Government. Title 17, U.S.C., §101 defines a U.S. Government work as a work prepared by a military Service member or employee of the U.S. Government as part of that person's official duties.

## References

1. Prostate Cancer - Cancer Stat Facts. Available at: https://seer.cancer.gov/statfacts/html/prost.html. (Accessed: 22nd August 2018)
2. Epstein, J. I. *et al.* A Contemporary Prostate Cancer Grading System: A Validated Alternative to the Gleason Score. *Eur. Urol.* **69,** 428–435 (2016).
3. Epstein, J. I., Allsbrook, W. C., Amin, M. B. & Egevad, L. L. The 2005 International Society of Urological Pathology (ISUP) Consensus Conference on Gleason Grading of Prostatic Carcinoma. *Am. J. Surg. Pathol.* **29,** 1228–1242 (2005).
4. Epstein, J. I. *et al.* The 2014 International Society of Urological Pathology (ISUP) Consensus Conference on Gleason Grading of Prostatic Carcinoma: Definition of Grading Patterns and Proposal for a New Grading System. *Am. J. Surg. Pathol.* **40,** 244–252 (2016).
5. NCCN Clinical Practice Guidelines in Oncology. Available at: https://www.nccn.org/professionals/physician_gls/default.aspx#prostate. (Accessed: 14th August 2018)
6. Persson, J. *et al.* Interobserver variability in the pathological assessment of radical prostatectomy specimens: findings of the Laparoscopic Prostatectomy Robot Open (LAPPRO) study. *Scand. J. Urol.* **48,** 160–167 (2014).
7. Veloso, S. G. *et al.* Interobserver agreement of Gleason score and modified Gleason score in needle biopsy and in surgical specimen of prostate cancer. *Int. Braz J Urol* **33,** 639–46; discussion 647–51 (2007).
8. Montironi, R., Lopez-Beltran, A., Cheng, L., Montorsi, F. & Scarpelli, M. Central prostate pathology review: should it be mandatory? *Eur. Urol.* **64,** 199–201; discussion 202–3 (2013).

9. Bottke, D. *et al.* Phase 3 study of adjuvant radiotherapy versus wait and see in pT3 prostate cancer: impact of pathology review on analysis. *Eur. Urol.* **64,** 193–198 (2013).
10. Egevad, L. *et al.* Standardization of Gleason grading among 337 European pathologists. *Histopathology* **62,** 247–256 (2013).
11. Netto, G. J., Eisenberger, M., Epstein, J. I. & TAX 3501 Trial Investigators. Interobserver variability in histologic evaluation of radical prostatectomy between central and local pathologists: findings of TAX 3501 multinational clinical trial. *Urology* **77,** 1155–1160 (2011).
12. Allsbrook, W. C., Jr *et al.* Interobserver reproducibility of Gleason grading of prostatic carcinoma: urologic pathologists. *Hum. Pathol.* **32,** 74–80 (2001).
13. Allsbrook, W. C., Jr *et al.* Interobserver reproducibility of Gleason grading of prostatic carcinoma: general pathologist. *Hum. Pathol.* **32,** 81–88 (2001).
14. Mikami, Y. *et al.* Accuracy of gleason grading by practicing pathologists and the impact of education on improving agreement. *Hum. Pathol.* **34,** 658–665 (2003).
15. Esteva, A. *et al.* Dermatologist-level classification of skin cancer with deep neural networks. *Nature* **542,** 115–118 (2017).
16. Haenssle, H. A. *et al.* Man against machine: diagnostic performance of a deep learning convolutional neural network for dermoscopic melanoma recognition in comparison to 58 dermatologists. *Ann. Oncol.* **29,** 1836–1842 (2018).
17. Gulshan, V. *et al.* Development and Validation of a Deep Learning Algorithm for Detection of Diabetic Retinopathy in Retinal Fundus Photographs. *JAMA* **316,** 2402–2410 (2016).
18. Ting, D. S. W. *et al.* Development and Validation of a Deep Learning System for Diabetic Retinopathy and Related Eye Diseases Using Retinal Images From Multiethnic Populations With Diabetes. *JAMA* **318,** 2211–2223 (2017).
19. Burlina, P. M. *et al.* Automated Grading of Age-Related Macular Degeneration From Color

Fundus Images Using Deep Convolutional Neural Networks. *JAMA Ophthalmol.* **135,** 1170–1176 (2017).

20. De Fauw, J. *et al.* Clinically applicable deep learning for diagnosis and referral in retinal disease. *Nat. Med.* **24,** 1342–1350 (2018).
21. Kermany, D. S. *et al.* Identifying Medical Diagnoses and Treatable Diseases by Image-Based Deep Learning. *Cell* **172,** 1122–1131.e9 (2018).
22. Rajpurkar, P. *et al.* CheXNet: Radiologist-Level Pneumonia Detection on Chest X-Rays with Deep Learning. *arXiv [cs.CV]* (2017).
23. Chilamkurthy, S. *et al.* Deep learning algorithms for detection of critical findings in head CT scans: a retrospective study. *Lancet* (2018). doi:10.1016/S0140-6736(18)31645-3
24. Ehteshami Bejnordi, B. *et al.* Diagnostic Assessment of Deep Learning Algorithms for Detection of Lymph Node Metastases in Women With Breast Cancer. *JAMA* **318,** 2199–2210 (2017).
25. Liu, Y. *et al.* Detecting Cancer Metastases on Gigapixel Pathology Images. *arXiv [cs.CV]* (2017).
26. Campanella, G., Silva, V. W. K. & Fuchs, T. J. Terabyte-scale Deep Multiple Instance Learning for Classification and Localization in Pathology. *arXiv [cs.CV]* (2018).
27. Arvaniti, E. *et al.* Automated Gleason grading of prostate cancer tissue microarrays via deep learning. *Sci. Rep.* **8,** 12054 (2018).
28. Steiner, D. F. *et al.* Impact of Deep Learning Assistance on the Histopathologic Review of Lymph Nodes for Metastatic Breast Cancer. *Am. J. Surg. Pathol.* (2018). doi:10.1097/PAS.0000000000001151
29. Liu, Y. *et al.* Artificial Intelligence–Based Breast Cancer Nodal Metastasis Detection. *Arch. Pathol. Lab. Med.* (2018). doi:10.5858/arpa.2018-0147-oa
30. van der Kwast, T. H. *et al.* Impact of pathology review of stage and margin status of radical

prostatectomy specimens (EORTC trial 22911). *Virchows Arch.* **449,** 428–434 (2006).

31. Srigley, J. R. *et al.* Protocol for the examination of specimens from patients with carcinoma of the prostate gland. *Arch. Pathol. Lab. Med.* **133,** 1568–1576 (2009).

32. Humphrey, P. A., Moch, H., Cubilla, A. L., Ulbright, T. M. & Reuter, V. E. The 2016 WHO Classification of Tumours of the Urinary System and Male Genital Organs—Part B: Prostate and Bladder Tumours. *Eur. Urol.* **70,** 106–119 (2016).

33. Epstein, J. I., Amin, M. B., Reuter, V. E. & Humphrey, P. A. Contemporary Gleason Grading of Prostatic Carcinoma: An Update With Discussion on Practical Issues to Implement the 2014 International Society of Urological Pathology (ISUP) Consensus Conference on Gleason Grading of Prostatic Carcinoma. *Am. J. Surg. Pathol.* **41,** e1–e7 (2017).

34. Sauter, G. *et al.* Clinical Utility of Quantitative Gleason Grading in Prostate Biopsies and Prostatectomy Specimens. *Eur. Urol.* **69,** 592–598 (2016).

35. Cox, D. R. Regression Models and Life-Tables. in *Springer Series in Statistics* 527–541 (1992).

36. Brimo, F., Schultz, L. & Epstein, J. I. The value of mandatory second opinion pathology review of prostate needle biopsy interpretation before radical prostatectomy. *J. Urol.* **184,** 126–130 (2010).

37. Zhou, M. *et al.* Diagnosis of 'Poorly Formed Glands' Gleason Pattern 4 Prostatic Adenocarcinoma on Needle Biopsy: An Interobserver Reproducibility Study Among Urologic Pathologists With Recommendations. *Am. J. Surg. Pathol.* **39,** 1331–1339 (2015).

38. Shah, R. B. *et al.* Diagnosis of Gleason Pattern 5 Prostate Adenocarcinoma on Core Needle Biopsy. *Am. J. Surg. Pathol.* **39,** 1242–1249 (2015).

39. Gordetsky, J. & Epstein, J. Grading of prostatic adenocarcinoma: current state and prognostic implications. *Diagn. Pathol.* **11,** 25 (2016).

40. Aeffner, F. *et al.* The Gold Standard Paradox in Digital Image Analysis: Manual Versus

Automated Scoring as Ground Truth. *Arch. Pathol. Lab. Med.* **141,** 1267–1275 (2017).

41. Wang, D., Khosla, A., Gargeya, R., Irshad, H. & Beck, A. H. Deep Learning for Identifying Metastatic Breast Cancer. *arXiv [q-bio.QM]* (2016).
42. Ehteshami Bejnordi, B. *et al.* Deep learning-based assessment of tumor-associated stroma for diagnosing breast cancer in histopathology images. in *2017 IEEE 14th International Symposium on Biomedical Imaging (ISBI 2017)* (2017). doi:10.1109/isbi.2017.7950668
43. Weinstein, J. N. *et al.* The Cancer Genome Atlas Pan-Cancer analysis project. *Nat. Genet.* **45,** 1113–1120 (2013).
44. Liu, J. *et al.* An Integrated TCGA Pan-Cancer Clinical Data Resource to Drive High-Quality Survival Outcome Analytics. *Cell* **173,** 400–416.e11 (2018).
45. Stephenson, A. J. *et al.* Defining biochemical recurrence of prostate cancer after radical prostatectomy: a proposal for a standardized definition. *J. Clin. Oncol.* **24,** 3973–3978 (2006).
46. Szegedy, C., Vanhoucke, V., Ioffe, S., Shlens, J. & Wojna, Z. Rethinking the Inception Architecture for Computer Vision. in *2016 IEEE Conference on Computer Vision and Pattern Recognition (CVPR)* (2016). doi:10.1109/cvpr.2016.308
47. Epstein, J. I. *et al.* A Contemporary Prostate Cancer Grading System: A Validated Alternative to the Gleason Score. *Eur. Urol.* **69,** 428–435 (2016).
48. Cohen, J. A Coefficient of Agreement for Nominal Scales. *Educ. Psychol. Meas.* **20,** 37–46 (1960).

## Supplementary Information

**Supplementary Table 1: Characteristics of the training and tuning datasets.**

| | | TCGA | Tertiary Teaching Hospital | Medical Laboratory | Total (%) |
|---|---|---|---|---|---|
| Number of patients | | 178 | 204 | 7 | 389 |
| Number of patients excluded due to non-gradable prostate cancer variants, extensive artifacts, or poor staining | | 43 | 4 | 0 | 47 |
| Number of patients included in the study | | 135 | 200 | 7 | 342 |
| Number of slides | | 170 | 1,016 | 40 | 1,226 |
| With slide-level Gleason scores | Number of slides | 144 | 988 | 27 | 1,159 (100%) |
| | GG 1 (slides) | 18 | 558 | 0 | 576 (50%) |
| | GG 2 (slides) | 32 | 218 | 0 | 250 (22%) |
| | GG 3 (slides) | 21 | 84 | 0 | 105 (9%) |
| | GG 4-5 (slides) | 73 | 128 | 27 | 228 (20%) |
| | GG 4 | 12 | 25 | 3 | 40 (3%) |
| | GG 5 | 61 | 103 | 24 | 188 (16%) |
| With region-level Gleason pattern annotations | Number of slides | 148 | 751 | 13 | 912 |
| | Number of patches | 14,422,449 | 97,737,450 | 455,947 | 112,615,846 (100%) |
| | Benign (patches) | 11,188,435 | 93,691,585 | 364,838 | 105,244,858 (93%) |
| | GP3 (patches) | 1,335,165 | 2,131,666 | 777 | 3,467,608 (3%) |
| | GP4 (patches) | 1,898,849 | 1,210,348 | 67,346 | 3,176,543 (3%) |
| | GP5 (patches) | 714,666 | 703,851 | 22,986 | 1,441,503 (1%) |

**Supplementary Table 1: Characteristics of the train and tuning datasets.** In these datasets, 1-7 slides from each patient were used, and each slide was reviewed by 3-5 pathologists. Slides were excluded from training/tuning if any pathologist deemed the slide ungradable due to variants or poor image quality. Slide-level Gleason scores and region-level Gleason pattern annotations were collected for overlapping subsets of these slides, with the breakdown described in the table above.

## Supplementary Table 2: Description of handling of variants.

| Prostate Cancer Variant | Action |
|---|---|
| Small Cell Carcinoma | Excluded |
| Mucinous prostatic adenocarcinoma | Excluded |
| Adenocarcinoma with signet ring cell like features | Graded via ISUP 2014 recommendations[1] |
| Prostate ductal adenocarcinoma | Excluded |
| Basal cell carcinoma | Excluded |
| **Histological Variant of Acinar Prostatic Adenocarcinoma** | **Action** |
| Mucinous fibroplasia | Graded via ISUP 2014 recommendations[1] |
| Foamy gland carcinoma | Graded via ISUP 2014 recommendations[1] |
| Paneth cell-like neuroendocrine differentiation. | Excluded |
| Treated prostatic adenocarcinoma | Excluded |
| Pseudohyperplastic prostatic adenocarcinoma | Graded via ISUP 2014 recommendations[1] |
| Intraductal carcinoma of the prostate | When found in conjunction with Gleason Gradable tumor, only the Gleason gradable component is graded (consistent with ISUP 2014 recommendations)[1] |

**Supplementary Table 2: Description of handling of prostate cancer variants and acinar adenocarcinoma histological variants.** Slides containing cancer variants and histological variants that are not Gleason gradable were excluded from the study (with the exception of intraductal carcinoma). Other variants are graded in a manner consistent with ISUP 2014 recommendations.

**Supplementary Table 3: Analysis on slides excluded from validation set.**

| Slide | Rationale for lack of confidence in diagnosis | Specialist 1 GG | Specialist 2 GG | Specialist 3 GG | DLS GG |
|---|---|---|---|---|---|
| 1 | need IHC - high grade tumor, but needs IHC to assess/quantify IDC vs pattern 5 | 4-5 | 4-5 | 4-5 | 4-5 |
| 2 | need IHC - 4+3 vs 4+5 (pattern 5 based on cribriform necrosis), but chatter artifact makes it difficult to tell; also would do IHC to r/o IDC vs pattern 4/5 areas | 3 | 3 | 3 | 4-5 |
| 3 | need IHC - high grade tumor case, but with areas of IDC vs pattern 4 vs pattern 4 with necrosis (pattern 5) | 3 | 4-5 | 3 | 4-5 |
| 4 | need IHC - areas of IDC vs pattern 4 vs pattern 4 with necrosis (pattern 5) | 2 | 3 | 2 | 2 |
| 5 | need IHC - likely 4+3 case, but given prominent areas of possible HGPIN/IDC need IHC to accurately quant pattern 4 | 3 | 3 | 3 | 3 |
| 6 | need IHC - given large areas of large cribriform glands (DDx HGPIN/IDC vs pattern 4), need IHC to accurately quantitate and grade | 4-5 | 4-5 | 4-5 | 4-5 |
| 7 | need IHC - large areas of possible IDC; need IHC to r/o vs pattern 4 and for accurate pattern 4/tumor vol % | 3 | 3 | 3 | 3 |
| 8 | need IHC - focal area of large cribriform glands present, would do both stains (r/o IDC vs pattern 4) and also levels as there may be necrosis (IDC vs pattern 5) | 2 | 2 | 2 | 2 |
| 9 | need IHC - definite invasive cancer present, but adjacent large cribriform glands with DDx of pattern 4 vs HGPIN needs IHCs to assess/quantitate | 3 | 3 | 3 | 3 |
| 10 | need IHC - areas of large cribriform glands needing IHC to eval IDC vs pattern 4 | 3 | 3 | 3 | 3 |
| 11 | need IHC - areas of definite large crib irregular glands of pattern 4, but some areas of probable IDC also (need IHC to accurately quantitate) | 4-5 | 4-5 | 4-5 | 4-5 |
| 12 | Needs another expert review. There is a pattern 3 that is not recognized. I don't see a pattern 5. | 3 | 3 | 3 | 4-5 |
| 13 | I suggest this case go to another expert as this case has many patterns and is good to our criteria titrated | 3 | 4-5 | 3 | 4-5 |
| 14 | No agreement among initial reviewers - suggest another expert opinion | 4-5 | 3 | 3 | 2 |
| 15 | Show to colleague(s) and order serial sections to confirm small % GG4 | 2 | 2 | 2 | 2 |
| 16 | Challenging slide - perhaps tissue was not well fixed as morphology was not great for grading. As such, I am not sure about GG5 - would show to a colleague(s). | 4-5 | 3 | 3 | 4-5 |
| 17 | Show to colleague(s) and order serial sections to confirm small % GG4 | 2 | 2 | 2 | 1 |
| 18 | Challenging case - would order serial sections to confirm minor GG5 (rule out tangential sectioning of GG4 poorly formed acini) as well as show to a colleague | 4-5 | 4-5 | 3 | 4-5 |
| 19 | require IHC | 3 | 3 | 2 | 3 |
| 20 | Would use basal cell IHC to rule in/rule out intraductal carcinoma | 3 | 3 | 2 | 4-5 |

**Supplementary Table 3: Analysis on slides excluded from validation set due to genitourinary specialist lack of confidence when diagnosing.** 20 slides were excluded from the analysis in the main text where the specialist adjudicator was not able to provide a confident diagnosis. Consults were subsequently provided by the other two GU experts. Of the 12 cases where the original adjudicator and two consulting experts came to a consensus, the DLS was concordant on 9 (highlighted in green) and within 1 grouping on the remaining 3 (highlighted in red).

## Supplementary Table 4: Comparison of DLS to pathologists' unadjusted accuracy.

| Grader | Unadjusted accuracy for grade group (95% CI) | p-value for comparison with DLS |
|---|---|---|
| Deep learning system | 0.698 (0.650, 0.746) | n/a |
| Mean among all 29 pathologists | 0.610 (0.563, 0.660) | 0.002 |
| Mean among 19-pathologist subgroup | 0.596 (0.529, 0.659) | <0.001 |
| Mean among 10-pathologist subgroup (A-J below) | 0.637 (0.588, 0.686) | 0.006 |
| Pathologist A | 0.526 (0.468, 0.577) | <0.001 |
| Pathologist B | 0.559 (0.502, 0.613) | <0.001 |
| Pathologist C | 0.592 (0.538, 0.644) | <0.001 |
| Pathologist D | 0.628 (0.574, 0.680) | 0.027 |
| Pathologist E | 0.647 (0.592, 0.695) | 0.16 |
| Pathologist F | 0.640 (0.589, 0.689) | 0.083 |
| Pathologist G | 0.668 (0.616, 0.716) | 0.40 |
| Pathologist H | 0.671 (0.616, 0.722) | 0.45 |
| Pathologist I | 0.716 (0.668, 0.764) | 0.59 |
| Pathologist J | 0.728 (0.683, 0.776) | 0.33 |

**Supplementary Table 4: Comparison of unadjusted concordance between the deep learning system, the cohort of 29 pathologists, and 10 individual pathologists (A-J).** The cohort of 29 pathologist comprised of 10 pathologists (A-J) that reviewed all 331 slides in the validation dataset and 19 pathologist that each reviewed a subset of the validation dataset. For the concordance of the individual 19 pathologists see Supplementary Table 5. Confidence intervals (CIs) were calculated with 1000 bootstrap replications. The statistical significance of the comparisons were performed using the permutation test.

## Supplementary Table 5: Comparisons of unadjusted accuracy on the validation set for the 19 pathologists who each reviewed a subset of the validation dataset.

| Grader | Number of slides in subset | Pathologist accuracy on subset (95% CI) | DLS accuracy on subset (95% CI) |
|---|---|---|---|
| Pathologist K | 62 | 0.306 (0.194, 0.435) | 0.742 (0.629, 0.840) |
| Pathologist L | 64 | 0.422 (0.312, 0.555) | 0.672 (0.570, 0.774) |
| Pathologist M | 55 | 0.545 (0.400, 0.655) | 0.618 (0.500, 0.746) |
| Pathologist N | 58 | 0.552 (0.414, 0.655) | 0.603 (0.483, 0.716) |
| Pathologist O | 54 | 0.556 (0.444, 0.648) | 0.630 (0.519, 0.759) |
| Pathologist P | 57 | 0.561 (0.439, 0.684) | 0.789 (0.675, 0.877) |
| Pathologist Q | 49 | 0.571 (0.438, 0.694) | 0.694 (0.592, 0.807) |
| Pathologist R | 40 | 0.575 (0.450, 0.725) | 0.700 (0.575, 0.850) |
| Pathologist S | 50 | 0.580 (0.440, 0.730) | 0.700 (0.600, 0.800) |
| Pathologist T | 50 | 0.580 (0.460, 0.690) | 0.700 (0.599, 0.850) |
| Pathologist U | 53 | 0.623 (0.500, 0.736) | 0.642 (0.528, 0.774) |
| Pathologist V | 49 | 0.633 (0.510, 0.755) | 0.673 (0.550, 0.786) |
| Pathologist W | 57 | 0.649 (0.509, 0.772) | 0.737 (0.622, 0.851) |
| Pathologist X | 60 | 0.650 (0.500, 0.750) | 0.700 (0.600, 0.800) |
| Pathologist Y | 46 | 0.674 (0.543, 0.783) | 0.652 (0.510, 0.761) |
| Pathologist Z | 44 | 0.682 (0.500, 0.818) | 0.705 (0.579, 0.818) |
| Pathologist AA | 50 | 0.700 (0.560, 0.820) | 0.700 (0.540, 0.830) |
| Pathologist AB | 41 | 0.732 (0.610, 0.854) | 0.732 (0.597, 0.878) |
| Pathologist AC | 53 | 0.736 (0.623, 0.887) | 0.698 (0.584, 0.821) |

**Supplementary Table 5**: **Comparisons of unadjusted accuracy on overlapping subsets of the validation set for the cohort of 19 pathologists.** Each pathologist reviewed a subset of the validation dataset, that collectively provided 3 annotations per slide for each of the 331 validation slides. In this subgroup analysis, the DLS's accuracy is greater than that of 14 of the 19 pathologists.

**Supplementary Table 6: Comparison of DLS to pathologists using other evaluation metrics.**

| Grader | Population-adjusted accuracy for grade group (95% CI) | p-value for comparison with DLS | Cohen's kappa for grade group (95% CI) | p-value for comparison with DLS | Accuracy for Gleason score (6-10) (95% CI) | p-value for comparison with DLS |
|---|---|---|---|---|---|---|
| Deep learning system | 0.720 (0.675, 0.762) | n/a | 0.585 (0.520, 0.651) | n/a | 0.770 (0.722, 0.813) | n/a |
| Mean among all 29 pathologists | 0.628 (0.578, 0.674) | <0.001 | 0.466 (0.398, 0.527) | 0.001 | 0.681 (0.638, 0.725) | <0.001 |
| Pathologist A | 0.515 (0.459, 0.569) | <0.001 | 0.365 (0.290, 0.430) | <0.001 | 0.672 (0.623, 0.723) | 0.002 |
| Pathologist B | 0.572 (0.519, 0.625) | <0.001 | 0.412 (0.341, 0.481) | <0.001 | 0.593 (0.540, 0.646) | <0.001 |
| Pathologist C | 0.615 (0.565, 0.660) | <0.001 | 0.457 (0.389, 0.522) | 0.001 | 0.703 (0.651, 0.752) | 0.039 |
| Pathologist D | 0.679 (0.635, 0.720) | 0.16 | 0.489 (0.415, 0.556) | 0.018 | 0.659 (0.607, 0.710) | <0.001 |
| Pathologist E | 0.603 (0.549, 0.655) | 0.003 | 0.506 (0.428, 0.573) | 0.10 | 0.734 (0.689, 0.777) | 0.27 |
| Pathologist F | 0.634 (0.581, 0.685) | 0.011 | 0.514 (0.441, 0.577) | 0.088 | 0.729 (0.683, 0.777) | 0.19 |
| Pathologist G | 0.656 (0.605, 0.712) | 0.070 | 0.530 (0.459, 0.600) | 0.21 | 0.734 (0.686, 0.782) | 0.25 |
| Pathologist H | 0.669 (0.618, 0.721) | 0.12 | 0.548 (0.475, 0.618) | 0.37 | 0.690 (0.638, 0.736) | 0.007 |
| Pathologist I | 0.727 (0.679, 0.775) | 0.81 | 0.613 (0.548, 0.678) | 0.45 | 0.769 (0.720, 0.815) | >.99 |
| Pathologist J | 0.758 (0.714, 0.801) | 0.18 | 0.622 (0.561, 0.690) | 0.33 | 0.773 (0.727, 0.818) | >.99 |

**Supplementary Table 6: Comparison of other evaluation metrics (adjusted accuracy for grade group, Cohen's Kappa for grade group, and accuracy for Gleason score) between the deep learning system (DLS), the cohort of 29 pathologists, and 10 individual pathologists (A-J).** The adjusted accuracy reflects a population-level GG distribution of 7397:8353:3106:1968.[2] Confidence intervals (CIs) were calculated with 1000 bootstrap replications. The statistical significance of the comparisons were performed using the permutation test.

## Supplementary Table 7: Comparison of %GP 3,4,5 quantitation.

| Grader | Mean absolute error for %GP3 (95% CI) | p-value for comparison with DLS | Mean absolute error for %GP4 (95% CI) | p-value for comparison with DLS | Mean absolute error for %GP5 (95% CI) | p-value for comparison with DLS |
|---|---|---|---|---|---|---|
| Deep learning system | 11.9 (10.0, 13.9) | n/a | 11.8 (10.5, 13.2) | n/a | 4.5 (3.4, 5.7) | n/a |
| Mean among all 29 pathologists | 16.0 (13.3, 18.7) | 0.004 | 17.8 (15.1, 20.7) | <0.001 | 5.2 (3.9, 6.6) | 0.26 |
| Pathologist A | 19.4 (16.9, 21.8) | <0.001 | 22.0 (19.3, 24.7) | <0.001 | 4.2 (3.0, 5.4) | 0.43 |
| Pathologist B | 19.5 (17.0, 22.3) | <0.001 | 22.6 (19.8, 25.5) | <0.001 | 5.4 (3.9, 6.9) | 0.19 |
| Pathologist C | 18.5 (15.9, 21.1) | <0.001 | 21.0 (18.3, 23.9) | <0.001 | 4.2 (3.1, 5.6) | 0.49 |
| Pathologist D | 13.1 (11.3, 14.9) | 0.36 | 15.8 (13.9, 17.7) | <0.001 | 5.0 (3.7, 6.5) | 0.50 |
| Pathologist E | 15.6 (13.7, 17.6) | 0.002 | 18.8 (16.8, 20.7) | <0.001 | 4.3 (3.3, 5.6) | 0.80 |
| Pathologist F | 15.2 (13.0, 17.4) | 0.002 | 17.0 (14.7, 19.3) | <0.001 | 4.9 (3.6, 6.2) | 0.51 |
| Pathologist G | 10.4 (9.1, 12.0) | 0.19 | 14.6 (12.8, 16.5) | 0.001 | 6.9 (5.5, 8.4) | <0.001 |
| Pathologist H | 10.2 (8.8, 11.7) | 0.12 | 14.5 (12.5, 16.3) | 0.003 | 6.5 (4.9, 8.2) | <0.001 |
| Pathologist I | 9.8 (8.3, 11.2) | 0.083 | 11.9 (10.3, 13.4) | >0.99 | 4.2 (3.2, 5.5) | 0.55 |
| Pathologist J | 10.2 (8.6, 11.8) | 0.13 | 12.2 (10.5, 14.0) | 0.68 | 3.9 (2.9, 5.0) | 0.10 |

**Supplementary Table 7: Comparison of Gleason pattern (GP) quantitation between the deep learning system (DLS), the cohort of 29 pathologists, and 10 individual pathologists.** Confidence intervals (CIs) were calculated with 1000 bootstrap replications. The statistical significance of the comparisons were performed using the permutation test.

## Supplementary Table 8: Comparison of %GP4 quantitation in GG2-3 slides and %GP5 quantitation in GG4-5 slides.

| Grader | Mean absolute error for %GP4 in GG 2-3 slides (95% CI) | p-value for comparison with DLS | Mean absolute error for %GP5 in GG 4-5 slides (95% CI) | p-value for comparison with DLS |
|---|---|---|---|---|
| Deep Learning System | 13.0 (11.5, 14.7) | n/a | 18.7 (14.3, 23.2) | n/a |
| Mean among all 29 pathologists | 20.5 (17.6, 24.0) | <0.001 | 22.0 (18.0, 26.9) | 0.30 |
| Pathologist A | 27.3 (23.8, 30.8) | <0.001 | 18.0 (13.5, 23.2) | 0.76 |
| Pathologist B | 25.1 (21.7, 28.7) | <0.001 | 24.7 (19.3, 30.4) | 0.076 |
| Pathologist C | 25.5 (22.1, 28.9) | <0.001 | 19.6 (14.0, 25.4) | 0.79 |
| Pathologist D | 19.1 (16.6, 21.7) | <0.001 | 24.2 (19.0, 29.6) | 0.12 |
| Pathologist E | 19.3 (17.0, 21.7) | <0.001 | 20.6 (16.1, 25.0) | 0.58 |
| Pathologist F | 18.0 (15.5, 20.6) | <0.001 | 19.0 (13.9, 24.3) | 0.89 |
| Pathologist G | 16.1 (14.0, 18.3) | 0.007 | 20.9 (15.7, 26.3) | 0.35 |
| Pathologist H | 15.4 (13.2, 17.6) | 0.044 | 24.0 (18.2, 31.0) | 0.046 |
| Pathologist I | 13.9 (12.0, 15.8) | 0.38 | 17.5 (12.8, 22.7) | 0.49 |
| Pathologist J | 14.6 (12.6, 16.8) | 0.12 | 17.4 (13.1, 22.0) | 0.53 |

**Supplementary Table 8: Comparison of Gleason pattern (GP) in Grade Groups (GG) 2-3 and 4-5 between the deep learning system (DLS), the cohort of 29 pathologists, and 10 individual pathologists (A-J).** Confidence intervals (CIs) were calculated with 1000 bootstrap replications. The statistical significance of the comparisons were performed using the permutation test.

## Supplementary Table 9: Sensitivity analysis excluding consult cases.

| Grader | Number of Slides excluded due to indication of a non-confident diagnosis | Pathologist's accuracy excluding consult cases (95% CI) | DLS accuracy excluding the same cases (95% CI) | p-value for comparison with DLS |
|---|---|---|---|---|
| Pathologist A | 3 | 52.0 (46.7, 57.3) | 69.2 (64.2, 74.1) | <0.001 |
| Pathologist B | 6 | 56.2 (50.6, 61.8) | 69.6 (64.8, 74.7) | <0.001 |
| Pathologist C | 2 | 59.9 (54.2, 65.3) | 69.4 (64.7, 74.3) | 0.002 |
| Pathologist D | 10 | 63.3 (58.0, 68.5) | 69.8 (65.0, 74.4) | 0.048 |
| Pathologist E | 3 | 64.7 (59.7, 69.7) | 69.3 (64.5, 74.4) | 0.22 |
| Pathologist F | 7 | 63.7 (58.4, 68.6) | 69.5 (64.7, 74.5) | 0.074 |
| Pathologist G | 2 | 67.0 (62.0, 72.2) | 69.5 (64.7, 74.4) | 0.50 |
| Pathologist H | 8 | 66.5 (61.5, 71.8) | 69.6 (64.8, 74.4) | 0.37 |
| Pathologist I | 9 | 71.4 (66.9, 76.0) | 69.6 (64.5, 74.4) | 0.60 |
| Pathologist J | 1 | 73.8 (69.1, 78.6) | 69.5 (64.6, 74.6) | 0.17 |

**Supplementary Table 9: Comparison between pathologists and DLS on Gleason scoring excluding slides indicated by pathologists as non-confident diagnosis.** The results are qualitatively similar to the results in Supplementary Table 4 with no material differences. Confidence intervals (CIs) were calculated with 1000 bootstrap replications. The statistical significance of the comparisons were performed using the permutation test.

**Supplementary Table 10: Adverse Clinical Event Models Derived From Gleason Pattern Quantitation and Fine-Grained Gleason Pattern Quantitation.**

| Source of Gleason pattern quantitation | Input features to Cox regression model describing tumor composition: (all based on % Gleason pattern) | C-index (95% CI) |
|---|---|---|
| Cohort-of-29 general pathologists | 3, 4, 5 | 0.674 (0.564, 0.782) |
| Genitourinary specialist pathologists | 3, 4, 5 | 0.690 (0.582, 0.800) |
| DLS | 3, 4, 5 | 0.697 (0.579, 0.790) |
| DLS | 3, 3.5, 4, 5 | 0.704 (0.586, 0.814) |
| DLS | 3, 3.5, 4, 4.5, 5 | 0.702 (0.577,0.812) |

**Supplementary Table 10: Comparison of Cox models for adverse clinical events (progression/biochemical recurrence) trained directly on quantified Gleason patterns and fine-grained Gleason Patterns.** Cox proportional hazards regression models were trained and evaluated on the validation set (n=331 slides), with Gleason patterns quantitation as input features. Features were provided by the cohort-of-29 pathologists, genitourinary specialists comprising the reference standard, and the DLS. As proof-of-concept, Cox models were also trained with additional features that provide finer-grained representations of tumor differentiation (see “Fine-grained Gleason Pattern” in Supplementary Methods). Confidence intervals (CI) were calculated via bootstrapping, and the median concordance index is presented for the cohort-of-29 pathologists (see Supplementary Methods).

## Supplementary Fig. 1: Confusion Matrices for the DLS and two pathologist subgroups

Specialist-Adjudicated Grade Group

| Deep Learning System's Grade Group (n=331 slides) | GG1 | GG2 | GG3 | GG4-5 |
|---|---|---|---|---|
| GG1 (n=77 slides) | 90% | 10% | 0% | 0% |
| GG2 (n=134 slides) | 22% | 66% | 10% | 1% |
| GG3 (n=62 slides) | 3% | 21% | 34% | 42% |
| GG4-5 (n=58 slides) | 2% | 2% | 7% | 89% |

| Subgroup of 10 pathologists (n=331 slides with 10 annotations/slide) | GG1 | GG2 | GG3 | GG4-5 |
|---|---|---|---|---|
| GG1 (n=77 slides) | 73% | 23% | 3% | 2% |
| GG2 (n=134 slides) | 20% | 58% | 16% | 5% |
| GG3 (n=62 slides) | 1% | 18% | 49% | 32% |
| GG4-5 (n=58 slides) | 1% | 4% | 16% | 79% |

| Subgroup of 19 pathologists (n=331 slides with 3 annotations/slide) | GG1 | GG2 | GG3 | GG4-5 |
|---|---|---|---|---|
| GG1 (n=77 slides) | 79% | 15% | 5% | 1% |
| GG2 (n=134 slides) | 31% | 51% | 14% | 4% |
| GG3 (n=62 slides) | 8% | 33% | 38% | 21% |
| GG4-5 (n=58 slides) | 1% | 6% | 21% | 72% |

**Supplementary Fig. 1: Confusion matrices highlighting the distribution of errors made by the DLS and two pathologist subcohorts.** The DLS is compared to the subgroup of 10 pathologists where each pathologist individually annotated every validation set slide, as well as the subgroup of 19 pathologists that collectively provided 3 reviews for every slide. The DLS shows greater accuracy in classifying slides as GG1, GG2, and GG4-5, and lower accuracy in classification of GG3 on the validation set as compared to these cohorts.

**Supplementary Fig. 2: Model and pathologist concordance with mixed grade labels.**

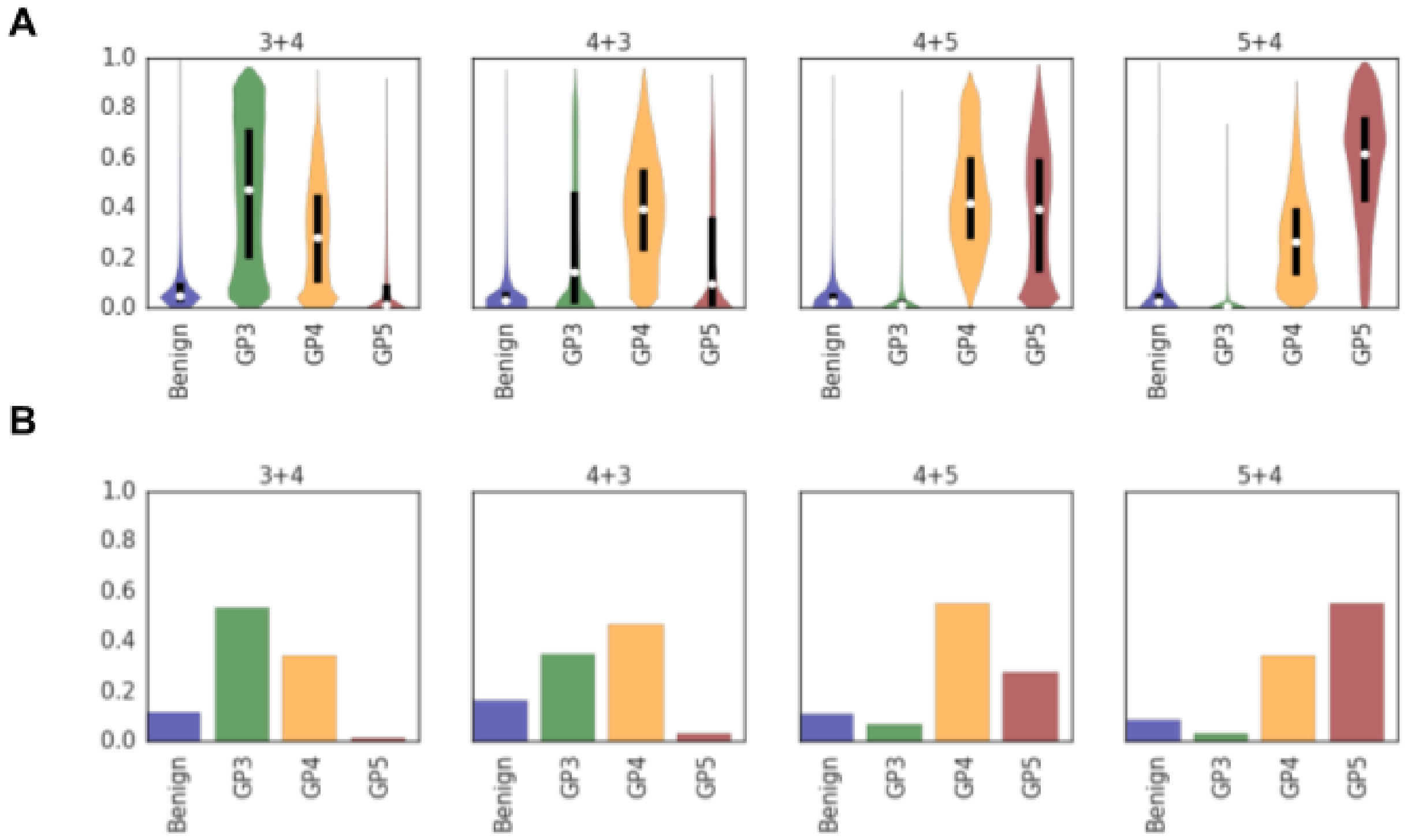

**Supplementary Fig. 2: Model and pathologist concordance with mixed grade labels.** When pathologists could not assign a single Gleason pattern to a region, they were instructed to assign a mixed grade label. Available mixed grade labels were '3+4', '4+3', '4+5', and '5+4'. These indicate that a region exhibits histological patterns characteristic of both Gleason patterns at the level of glands, and they are an extension to the Gleason grading system which allow humans to represent a small slice of the continuum of Gleason grading. To further investigate the deep learning system's ability to quantitatively represent the ambiguities present in the Gleason grading system, we examine the model's output in those cases in which a pathologist provided a mixed grade. **A,** Distributions of predicted likelihood of each GP by the DLS on patches labeled as a mixed grade by at least one pathologist. The DLS represents "in-between" patterns by exhibiting mixed likelihood between multiple labels. **B,** The distribution of other pathologist grades for those patches which were given a mixed grade by at least one pathologist.

## Supplementary Fig. 3: Extended visualization of Gleason patterns.

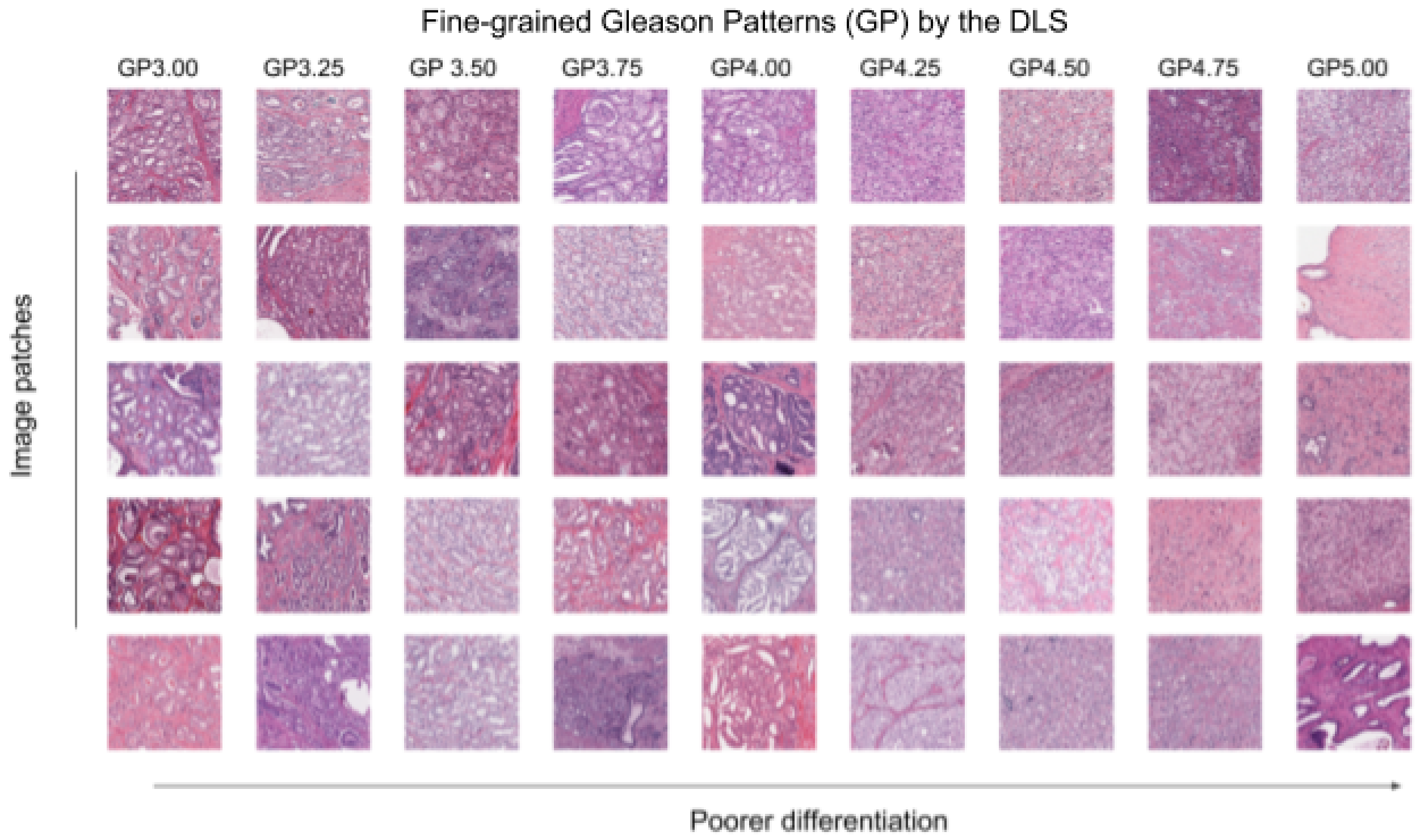

**Supplementary Fig. 3: Extended visualization of Gleason patterns.** The continuum of prostate cancer Gleason Patterns (GP) learned by the DLS reveals finer categorization of the well-to-poorly differentiated spectrum. The top row highlights the DLS GP categorization followed by H&E images that are predicted to be the corresponding quantitative GP. Columns 1, 5, and 9 represent 100% confidence in GP 3, 4, and 5 respectively. The columns in between represent quantitative GPs that are in between these defined categories.

**Supplementary Fig. 4: Screenshot of the tool used for region-level annotations.**

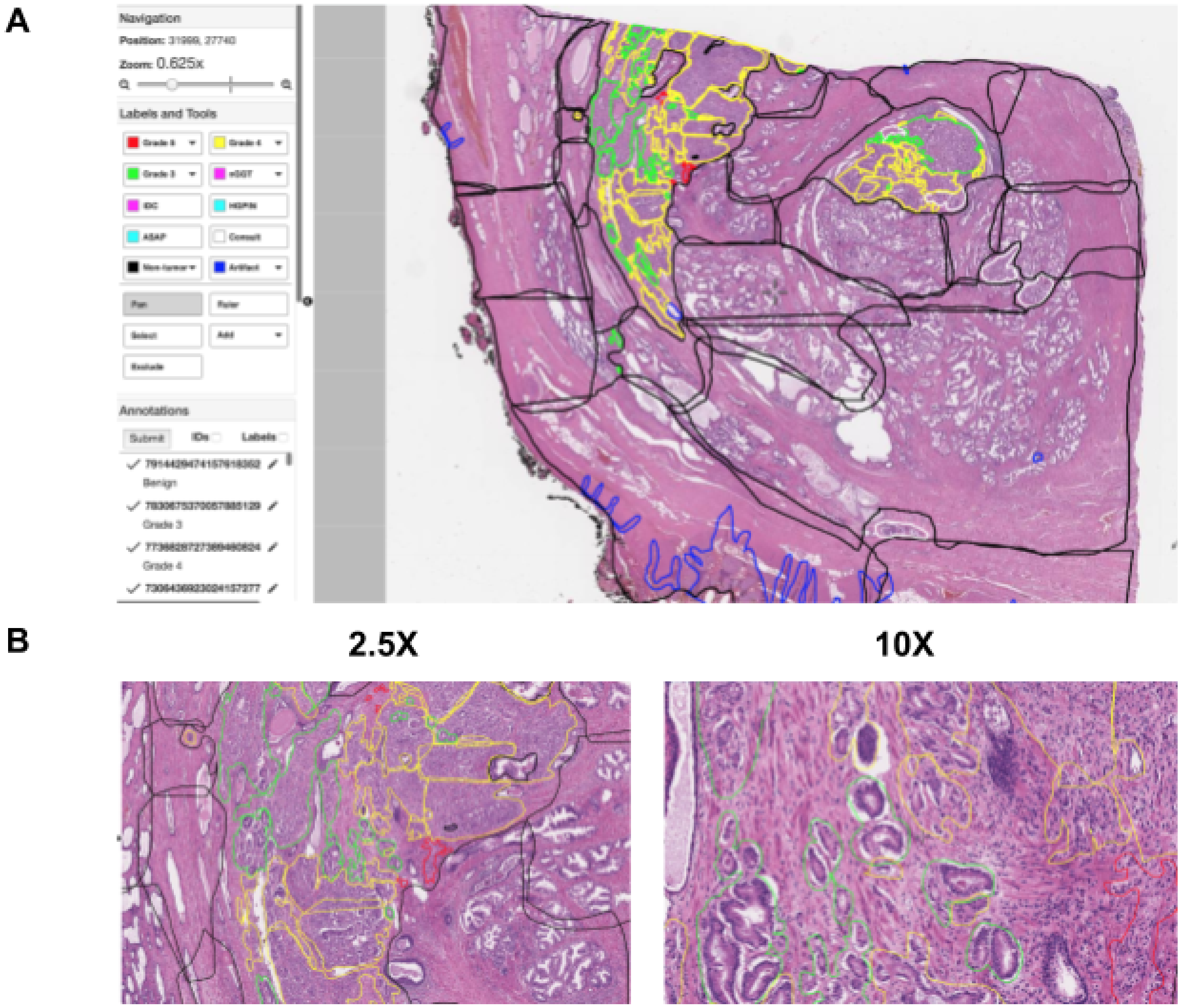

**Supplementary Fig. 4: Screenshots of the tool used for region-level annotations. A,** An overview of the tool zoomed out to 0.625X. A user annotates a region by first selecting a label category on the left and then outlining the corresponding regions direct on the slide. This custom free-hand drawing tool also has the ability to zoom between different objective powers as appropriate. **B,** Screenshots of annotations on tissue regions at additional magnifications: 2.5X and 10X. Most annotations were done between 5-20X.

Whole Slide Images

| | Training | Tuning | Validation |
|---|---|---|---|
| Region-Level annotations | 1 pathologist per slide | 3 pathologists per slide:<br><br>Image patches where ≥2 pathologists were concordant were used | 3 pathologists per slide:<br><br>Analysis on patches where all 3 pathologists provided a categorization |
| Slide-Level annotations | ≥3 pathologist reviews:<br><br>Most common grade with ties in favor of the more severe grade for sensitivity | ≥3 pathologist initial reviews, and final grade provided by a genitourinary expert | ≥3 pathologist initial reviews, and final grade provided by a genitourinary expert |

**Supplementary Fig. 5: Development of datasets used for training, tuning, and validation.** Region-level datasets and the slide-level training datasets were provided by pathologists, while the generation of the slide-level tuning and validation datasets involved genitourinary expert pathologists. More details can be found in the Grading sections of the Methods and the Supplementary Methods.

# Supplementary Methods

## Grading

### Pathologist Slide-Level Gleason Scoring Protocol

Slides used for training were reviewed by at least 3 and up to 7 pathologists (median 4). The label for each slide was determined by the most common annotation provided by the pathologists, while breaking ties in favor of the more severe grade to encourage higher DLS sensitivity. Tuning slides were initially reviewed by 3 to 5 pathologists and subsequently adjudicated by 1 of 3 genitourinary specialists (similar to the validation dataset).

We derived the slide-level Gleason score (e.g. 3+4) from the predominant GP and next-most-common GP. This is used instead of the directly provided Gleason scores because we noted inconsistent application of tertiary replacement (replacing the secondary Gleason score with '5' if %GP5 is greater than 5%), leading to even greater diagnostic variability.[2] The GG (e.g. GG2) was then directly determined using the Gleason score using the published definitions.[2] Pathologists were additionally instructed to note if a slide contained histologic variants (listed in Supplementary Table 2), did not contain tumor, or if they were not confident in their diagnosis.

### Pathologist Region-Level Annotation Protocol

The region-annotations for all datasets (training, tuning, and validation) were performed using custom free-hand drawing tools in a custom histopathology viewer (see Supplementary Fig. 4) with the ability to zoom between magnifications. Most annotations were performed between 5X and 20X magnifications. Artifacts that affected the ability to make a confident interpretation were labeled as artifacts, and regions where the pathologists were not able to assign confident categorizations based on their best clinical judgement were assigned a "consult" label. Regions where different GPs were either ambiguous or difficult to delineate exactly were assigned mixed-grade labels such as '3+4'. Perineural and lymphovascular invasive tumor and intraductal carcinoma were labeled as non-Gleason-gradable tumors.

For the training slides, at least one pathologist non-exhaustively annotated characteristic regions of each slide (annotated tissue for each slide <1% to 100%, median of 57%). For the tuning slides, we obtained higher-confidence labels by asking three pathologists for exhaustive annotations. In this set, to improve annotation efficiency (retaining slide-level diversity while reducing the overall annotation workload), the pathologists annotated only a subset of each slide, specifically two 3.8x3.8mm square regions from each quadrant on the slide. The locations of the two squares within each quadrant were randomly selected, and all three pathologists annotated the same eight regions (annotated tissue for each slide <1% to 35%, median of 14%). Only image patches with concordance between at least two annotators were used.

To train the stage-1 DLS, we processed the training dataset annotations to retain only regions with unambiguous labels. Ambiguity arising from multiple different labels were resolved by majority vote. Regions labeled ‘artifact’ were interpreted as non-tumor to reduce false positive predictions on artifact-containing regions. Regions labeled as ‘mixed-grade’ were interpreted as the primary pattern (*e.g.*, ‘5+4’ was interpreted as GP5), based on empirical observations of a resultant boost in stage-1 region-level accuracy. For the tuning datasets, only regions for which all three annotators provided a label were considered (similar to the validation dataset). In the main text, we report results only for patches labeled non-tumor, GP3, GP4, GP5. The analysis of image patches that are labeled with mixed-grades are presented in Supplementary Fig. 2.

**Development of the Deep Learning System**

We used a Inception-V3[3] image classification network, with fewer filters per layer (depth_multiplier=0.1) and modified to be fully-convolutional to improve inference throughput on whole-slide images (manuscript under review). To avoid introducing grid artifacts, the fully-convolutional modification involved using ‘VALID’ instead of ‘SAME’ padding in convolutions and differential cropping of the output of ‘branches’ in the Inception architecture. This network takes as input image patches of size 911x911 pixels at 10X magnification (equivalent to 911 $\times$ 911 μm). The region “assessed” by the network is a 32 $\times$ 32 μm region centered in each image patch.

The training process involved feeding image patches into the network with a specific sampling strategy to avoid bias towards specific slides or classes: first select a class according to the ratios 4:2:2:1 for the four classes respectively, then select a slide containing regions labeled as that class, and finally select an image patch from that slide. To help improve generalization performance, we applied data augmentation techniques to randomly perturb the actual images seen by the neural network (image perturbations for saturation, contrast, brightness, hue, and orientation) during training.[4] Training was performed in TensorFlow[5] using an RMSProp optimizer[6] and the softmax cross-entropy loss function. Hyperparameters such as the four-class sampling ratios, magnitude of image perturbations, the learning rate decay schedule, and L2 regularization decay were tuned via Gaussian-Bandit search on *Google Vizier*.[7] After tuning model hyperparameters, hard negative mining and ensembling were employed to further improve model performance. See below section for details of hard-negative mining.

After model convergence (as determined by the patch-level four-way classification performance on the tuning set, as measured by Cohen's kappa), we applied ensembling at three levels. First, the actual network weights used were smoothed using an exponential moving average with decay constant of 0.9999. Second, for each patch, the model predictions across eight image orientations (4 90° rotations and 2 left-right flips) were averaged using the geometric mean. Lastly, these orientation-averaged predictions were again averaged across four independently trained models (each with a separate hard-negative mining process), again using the geometric mean.

In the second stage of the DLS, we first calibrated each region's class predicted likelihoods. The calibration weights were determined empirically to produce the best slide-level predictions on the tuning set. Next, to obtain a categorical prediction for each patch, we applied the argmax function. Finally, each slide's patch-level predictions were summarized as four features: %Tumor, %GP3, %GP4, and %GP5. We linearly rescaled these features to have a minimum of 0 and a maximum of 1 in the training set, and trained a k-nearest neighbor (kNN) model for each prediction task: 4-way GG classification (GG 1, 2, 3 or 4-5), and each of the three binary classifications of GG $\geq$ 2, GG $\geq$ 3, and GG $\geq$ 4. The hyperparameter "k" (number of nearest neighbors) and neighbor-weighting method (uniform versus reciprocal of distance) were selected based on the performance of each model on the tuning set, as measured by kappa for GG and area under receiver operating characteristic (AUC) for the binary predictions. Our final selected hyperparameters were k=24 with uniform neighbor weighting. In addition, we evaluated the performance of several other machine learning algorithms, such as logistic regression, and random forest on the tuning set. kNN was selected to avoid over-fitting based on the limited size of the slide-level dataset and for ease of interpretability (as visualized in Fig. 1).

**Hard-Negative Mining**

Our DLS stage-1 development process includes large scale, continuous “hard-negative mining” which aims to improve algorithm performance by running inference on the entire training dataset to isolate the hardest examples and further refine the algorithm using these examples.

In hard negative mining, inference was run hourly by applying the partially-trained network to the entire training dataset (over 112 million image patches) for the entire duration of the training. These inference results were then used to alter the patch-sampling probabilities for every slide in the training set. For a given class in each slide, these sampling probabilities were initialized at the start of training to be uniform across all image patches. After every inference round, the sampling probabilities were updated to be proportional to the cross-entropy loss of each patch, such that incorrect classifications were sampled more frequently. In other words, as training proceeded, the DLS learned from harder and harder examples, which improved its accuracy more efficiently than random examples. While previous works employing deep learning on histopathology images have employed hard negative mining in an offline “batch-mode”[8–10], we observed that performance improves with the frequency of inference on the entire training dataset, resulting in the “quasi-online” hard-negative mining approach (>30,000 DLS stage-1 inferences per second) used here. We anticipate that the benefits of this continuous hard negative mining approach may be applicable to developing other deep learning algorithms on histopathology images as well.

For histopathology applications on whole-slide imaging, hard negative mining is a computationally expensive process, requiring inference over 112 million image patches in our training dataset.

## Fine-grained Gleason Pattern (GP)

To provide a more quantitative GP that smoothly interpolates between existing GPs (3, 4, and 5), we processed the calibrated DLS-predicted likelihood for each GP. First, the predictions for the two GPs with highest confidences were used to interpolate between the two GPs using the formula $likelihood_1$ / ($likelihood_1$ + $likelihood_2$). For example, if the GP 3,4,5 predictions were [0.7, 0.2, 0.1], then the computed value was 0.7 / (0.7 + 0.2) = 0.78, and the quantitative GP was 3+0.78 = 3.78. To visualize these quantitative GPs (e.g. in Fig. 4a), we used the International Commission on Illumination “Lab” (CIELAB) color space, which is designed to be perceptually uniform with respect to the underlying numerical values. To select regions that represent desired quantitative GPs (Fig. 4c and Supplementary Fig. 3), we located the image patches among all validation dataset slides for which the computed quantitative GP most closely matched the desired GP (e.g. 3.5).

## Statistical Analysis

### Comparison with the Cohort-of-29

Comparison of the DLS with the cohort-of-29 pathologists required a modified permutation test[11] to account for the different numbers of slide-level annotations provided by each pathologist. Specifically, 10 pathologists annotated all the slides (331 annotations each), while 19 pathologists collectively annotated all the slides 3 times (about 50±10 annotated slides by each pathologist). The 10 pathologists that annotated all the slides were selected based on slide reviewing speed and availability. To represent each pathologist equally, we modify the permutation test as follows: define our test statistic as the difference between the DLS accuracy and the mean accuracy among pathologists in the cohort-of-29. In each iteration of the permutation test, for each slide, randomly swap the model's given rating with one of the 14 ratings given for that slide (allowing the model to "swap" with itself with probability 1/14), and compute the test statistic on the result. After 5000 iterations, this gives a null distribution of the test statistic against which we compare the observed difference to compute a two-tailed p value.

In the risk stratification analyses, the cohort-of-29 pathologists annotations were sampled to approximate equal representation of each pathologist. For each slide, the sampled annotation can come from either one of  subgroup-of-10 annotations or one of the 3 available subgroup-of-19 annotations. Specifically, for each slide, an annotation was selected from one of the 10 available subgroup-of-10 annotations with 1/29 probability, or from one of the 3 available subgroup-of-19 annotations with (19/29)*(1/3) probability.

### Bootstrap Approach for Confidence Intervals

To compute confidence intervals for the pools of 10, 19, and 29, we bootstrapped both slides and annotators by resampling both with replacement in each iteration of the bootstrap. In the case of the pool of 29, to replicate our experimental design in each iteration, we separately resampled the subsets of 10 and 19.

## Supplementary Results

### DLS Region-level Errors

Here, we present a qualitative analysis of the errors made by the DLS's first stage, at the region level. Several errors were related to spatial localization. For example, the spatial extent of each predicted Gleason pattern region was sometimes imprecise; if two tumor-containing regions were separated by a small strip of non-tumor tissue, the DLS would sometimes categorize the intervening non-tumor as tumor.

Similarly, delineating the precise stroma-tumor interface was difficult for the DLS, in particular for GP5 and stroma (non-tumor). This was likely because GP5 can present as individual tumor cells in a background of connective tissue, and outlining each individual cell was impractical. The "impurity" of the underlying region-level annotation made it difficult to develop a DLS that was precision with respect to the boundary.

In many other cases, the errors made by the DLS was one where the underlying histology was ambiguous, such as when a tangential cut into a GP3 region caused it to resemble the fused-gland pattern that defines GP4. Because the DLS was trained to interpret the image patch surrounding the region, it will not take into account context from beyond its input image.

The remaining region-level errors involved true prediction mistakes that will naturally improve with more data. The second stage of the DLS is fairly robust against all of these errors by summarizing the predictions from all regions on the slide as a small number of features.

## Supplementary References

1. Epstein, J. I. *et al.* The 2014 International Society of Urological Pathology (ISUP) Consensus Conference on Gleason Grading of Prostatic Carcinoma: Definition of Grading Patterns and Proposal for a New Grading System. *Am. J. Surg. Pathol.* **40,** 244–252 (2016).
2. Epstein, J. I. *et al.* A Contemporary Prostate Cancer Grading System: A Validated Alternative to the Gleason Score. *Eur. Urol.* **69,** 428–435 (2016).
3. Szegedy, C., Vanhoucke, V., Ioffe, S., Shlens, J. & Wojna, Z. Rethinking the Inception Architecture for Computer Vision. in *2016 IEEE Conference on Computer Vision and Pattern Recognition (CVPR)* (2016). doi:10.1109/cvpr.2016.308
4. Liu, Y. *et al.* Detecting Cancer Metastases on Gigapixel Pathology Images. *arXiv [cs.CV]* (2017).
5. Martin Abadi and Paul Barham and Jianmin Chen and Zhifeng Chen and Andy Davis and Jeffrey Dean and Matthieu Devin and Sanjay Ghemawat and Geoffrey Irving and Michael Isard and Manjunath Kudlur and Josh Levenberg and Rajat Monga and Sherry Moore and Derek G. Murray and Benoit Steiner and Paul Tucker and Vijay Vasudevan and Pete Warden and Martin Wicke and Yuan Yu and Xiaoqiang Zheng. TensorFlow: A System for Large-Scale Machine Learning. in (USENIX Association).
6. Geoffrey Hinton Nitsh Srivastava. Neural Networks for Machine Learning. *University of Toronto Computer Science* Available at: http://www.cs.toronto.edu/~tijmen/csc321/slides/lecture_slides_lec6.pdf. (Accessed: 14th August 2018)
7. Daniel Golovin, Benjamin Solnik, Subhodeep Moitra, Greg Kochanski, John Karro, D. Sculley. Google vizier: A service for black-box optimization. in *Proceedings of the 23rd ACM SIGKDD International Conference on Knowledge Discovery and Data Mining* 1487–1495 (Google, August 13 - 17, 2017).
8. Wang, D., Khosla, A., Gargeya, R., Irshad, H. & Beck, A. H. Deep Learning for Identifying Metastatic Breast Cancer. *arXiv [q-bio.QM]* (2016).
9. Ehteshami Bejnordi, B. *et al.* Diagnostic Assessment of Deep Learning Algorithms for Detection of

Lymph Node Metastases in Women With Breast Cancer. *JAMA* **318,** 2199–2210 (2017).

10. Ehteshami Bejnordi, B. *et al.* Deep learning-based assessment of tumor-associated stroma for diagnosing breast cancer in histopathology images. in *2017 IEEE 14th International Symposium on Biomedical Imaging (ISBI 2017)* (2017). doi:10.1109/isbi.2017.7950668
11. Chihara, L. M. & Hesterberg, T. C. *Mathematical Statistics with Resampling and R*. (2018).